\documentclass[11pt]{article}

\usepackage[final]{acl}

\usepackage{times}
\usepackage{latexsym}
\usepackage[T1]{fontenc}
\usepackage[utf8]{inputenc}
\usepackage{microtype}
\usepackage{inconsolata}
\usepackage{graphicx}
\usepackage{amsmath}
\usepackage{amssymb}
\usepackage{booktabs}
\usepackage{multirow}
\usepackage[dvipsnames,table]{xcolor}
\usepackage{subcaption}
\usepackage{enumitem}
\usepackage{pifont}
\usepackage[most]{tcolorbox}
\usepackage{algorithm}
\usepackage{algorithmic}
\usepackage{tikz}
\usepackage{pgfplots}
\pgfplotsset{compat=1.18}
\usepgfplotslibrary{polar}
\usepackage{tabularx}

\usepackage{alphalph}
\makeatletter
\AddToHook{cmd/appendix/after}{%
}
\makeatother

\newcommand{\cmark}{\ding{51}}
\newcommand{\xmark}{\ding{55}}
\newcommand{\cfr}{\textsc{CFR}}
\newcommand{\rcfr}{\textsc{Routed-CFR}}

\newtcolorbox{constraintbox}[1][]{
  colback=blue!3!white, colframe=blue!50!black,
  fonttitle=\bfseries\small, title={#1},
  boxrule=0.6pt, arc=2pt, left=3pt, right=3pt, top=2pt, bottom=2pt,
  fontupper=\small
}

\newtcolorbox{findingbox}[1][]{
  colback=teal!5!white, colframe=teal!75!black,
  fonttitle=\bfseries\small, title={#1},
  boxrule=0.6pt, arc=2pt, left=3pt, right=3pt, top=2pt, bottom=2pt,
  fontupper=\small
}

\title{Constraint-First Reasoning: A Training-Free Protocol\\for Exploiting Answer-Space Constraints\\in Mathematical Problem Solving}

\author{
{\mdseries Hongbo Ma$^{1,*}$, Bangji Yang$^{2,*}$, Yunqian Selina Cheng$^{1,*}$},\\
Jiajun Fan$^{2}$, Hanwen Zhang$^{1}$, Ge Liu$^{2,\dagger}$\\
$^{1}$Tsinghua University \qquad $^{2}$University of Illinois Urbana-Champaign\\
$^{*}$Equal contribution \qquad $^{\dagger}$Corresponding author
}

\begin{document}
\maketitle

\begin{abstract}
%==================================================================
Large language models can derive a plausible mathematical object yet still violate explicit requirements---for example, by omitting a modular reduction, returning a non-integer, or using the wrong encoded answer form. We introduce \textbf{Constraint-First Reasoning} (\cfr{}), a training-free two-stage prompting protocol: Stage~1 extracts and summarizes constraints entailed by the problem, and Stage~2 solves while checking intermediate and final results against that summary. \rcfr{} activates the two-stage protocol only when a text-only regex router detects restrictive cues; otherwise it uses direct chain-of-thought (CoT). Across AIME, CMIMC, BRUMO, and AIMO\_AMC, the method improves direct CoT on multiple backbones. We further report convention-controlled routing experiments, matched prompting baselines, problem-level paired tests, decoding robustness, constraint-quality audits, total-token accounting, and an OlympiadBench evaluation. These analyses position CFR as a targeted test-time intervention whose benefit depends on recoverable constraints and reliable Stage~1 extraction, rather than as a general-purpose replacement for mathematical reasoning.
\end{abstract}

%==================================================================
\section{Introduction}
\label{sec:intro}
%==================================================================

Competition mathematics problems embed constraints in natural language: ``find the remainder when $N$ is divided by 1000,'' ``how many positive integers satisfy\ldots,'' ``express your answer as $p+q$ where $p$ and $q$ are coprime.'' These constraints define the \emph{feasible answer region}---a small, structured subset of all possible values. A solver that ignores them risks producing answers that violate stated requirements despite correct underlying reasoning.

Why do constraint violations occur so frequently? We identify three root causes. First, \emph{attention dilution}: in long reasoning chains, early constraint information can fade before the model reaches a final answer~\citep{liu2023lost}. Second, \emph{implicit encoding}: models trained on solution traces handle constraints implicitly---they may notice a modular condition mid-solution or check integer constraints post-hoc---but this implicit handling breaks under distribution shift. Third, \emph{format ambiguity}: problems with encoded answer formats (``find $m+n$ where $\gcd(m,n)=1$'') require multi-step post-processing that models frequently skip or misapply.

\begin{figure}[t]
  \centering
  \includegraphics[width=\columnwidth]{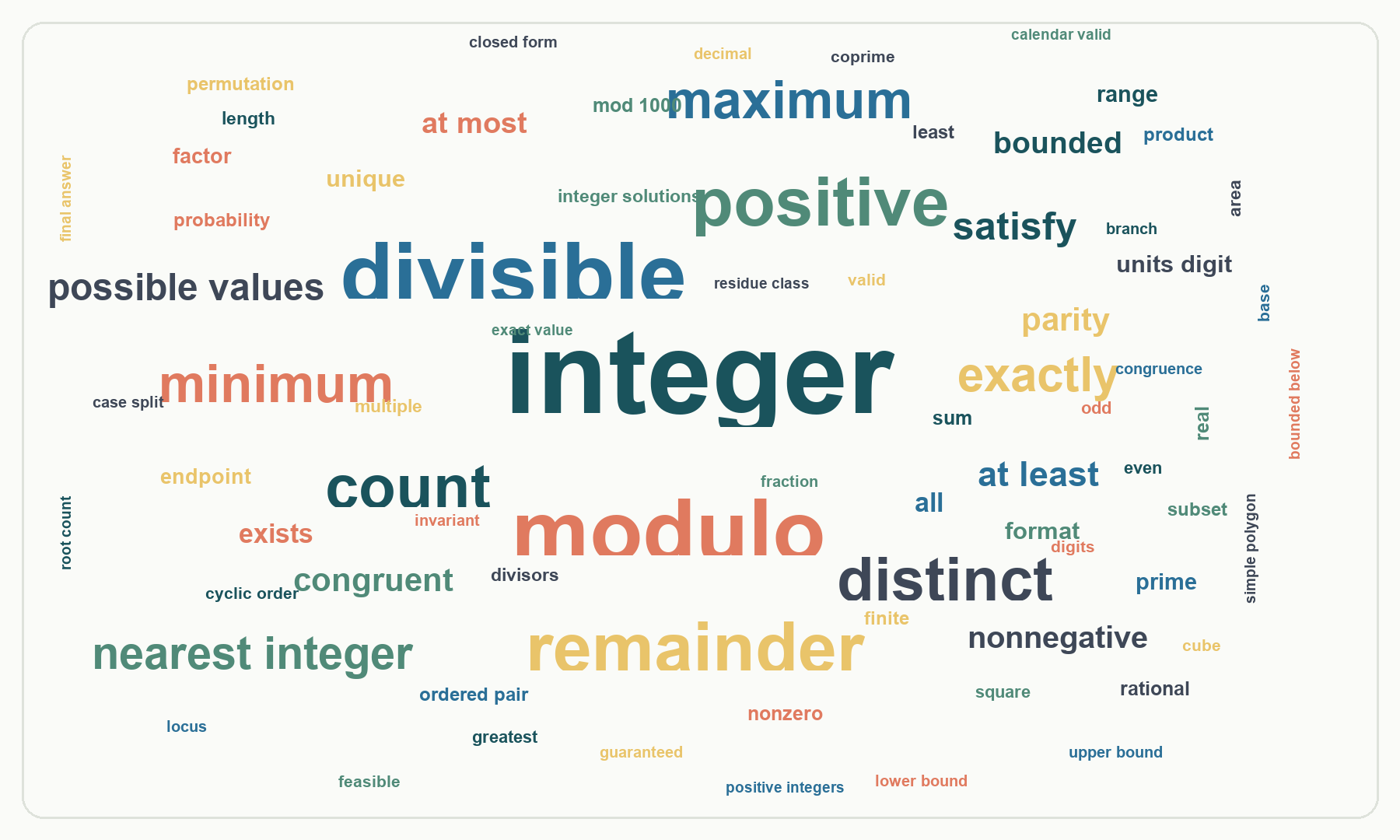}
  \caption{\textbf{Restrictive lexical cues in competition mathematics.} Larger words correspond to cue types emphasized by our router, including integer-answer formats, divisibility and modular conditions, bounds, extremal operators, exactness requirements, and counting restrictions. These cues motivate CFR: before solving, identify what any valid answer must satisfy.}
  \label{fig:constraint-wordcloud}
\end{figure}

Current reasoning models, trained to generate extended chains of thought~\citep{deepseekr1,openai2024o1}, handle constraints implicitly. A model may follow a plausible derivation while leaving answer restrictions implicit until the end; explicit constraint tracking is designed to keep those restrictions visible throughout solving.

We propose \textbf{Constraint-First Reasoning} (\cfr{}), a training-free protocol that makes constraint handling explicit and systematic. The key insight: \emph{extract constraints before solving, not after}. This mirrors human competition practice, where experienced contestants identify the required answer form before attempting a solution~\citep{polya1945solve}. Front-loading constraint identification ensures the solver maintains awareness of answer-space requirements throughout its reasoning chain.

\cfr{} operates in two stages:
\begin{enumerate}[nosep,leftmargin=*]
    \item \textbf{Constraint Extraction \& Prompted Summary}: Extract answer-space constraints (domain, modular, parity, bounds, format) and problem-structure constraints (e.g., invariants or branch conditions) from the problem statement, then summarize their implications for solving.
    \item \textbf{Constraint-Guided Solving}: Solve under the extracted constraints, checking intermediate results against the constraint specification at each major step.
\end{enumerate}

We further introduce \rcfr{}, which prepends a regex-based router that checks whether the problem text contains restrictive lexical cues. Problems without such cues bypass the pipeline and proceed directly to standard chain-of-thought reasoning. This purely syntactic router decides whether to use the full pipeline before generation begins.

Our experiments use Table~\ref{tab:model-performance-main} and Table~\ref{tab:prompting-main} as the primary empirical evidence. Detailed statistical, router, and error analyses are reported in the appendix. They reveal a clear and interpretable pattern:
\begin{itemize}[nosep,leftmargin=*]
    \item Under the text-only router, Table~\ref{tab:model-performance-main} shows positive AIME/CMIMC average gains for all four evaluated backbones; the magnitude varies with Stage~1 reliability.
    \item Table~\ref{tab:prompting-main} compares CFR with matched planning, verification, format-only, and token-matched prompting alternatives under a single evaluation protocol.
    \item The appendix reports pooled problem-level analyses after removing the AIME routing convention and correcting for multiple comparisons.
    \item Table~\ref{tab:prompting-main} makes the accuracy--cost trade-off explicit: routing reduces, but does not eliminate, the inference overhead of the two-stage protocol.
\end{itemize}

\subsection{Evaluation Protocol and Method Labels}
\label{sec:eval-protocol}

Unless explicitly stated otherwise, every accuracy in the main paper is \textbf{pass@1 averaged over four independent runs} (avg@4). Each run produces one answer per problem; we use the same decoding setting, answer extractor, and grading script for all methods in a comparison. We do not use self-consistency, majority voting, best-of-$k$ selection, or answer-format filtering in these primary results. For paired analyses, the four pass@1 outcomes are first averaged within each problem, so the problem---not an individual generation---is the unit of inference. Appendix~\ref{app:prompting-comparison} retains a clearly labeled pass@4 exploratory comparison, which is not pooled with or used to substantiate the main pass@1 claims.

We use \textbf{Routed-CFR} for the text-only router followed by the two-stage pipeline, and \textbf{CFR (always-on)} for the same two-stage pipeline without routing. This distinction is used consistently in tables and captions. ``CFR'' without a qualifier refers only to the general constraint-first protocol, not to an ambiguous experimental variant.

These findings position \cfr{} not as a universal reasoning enhancer but as a targeted intervention for settings where answer-space constraints are informative and the backbone can use them.

\begin{figure*}[t]
\centering
\includegraphics[width=\textwidth]{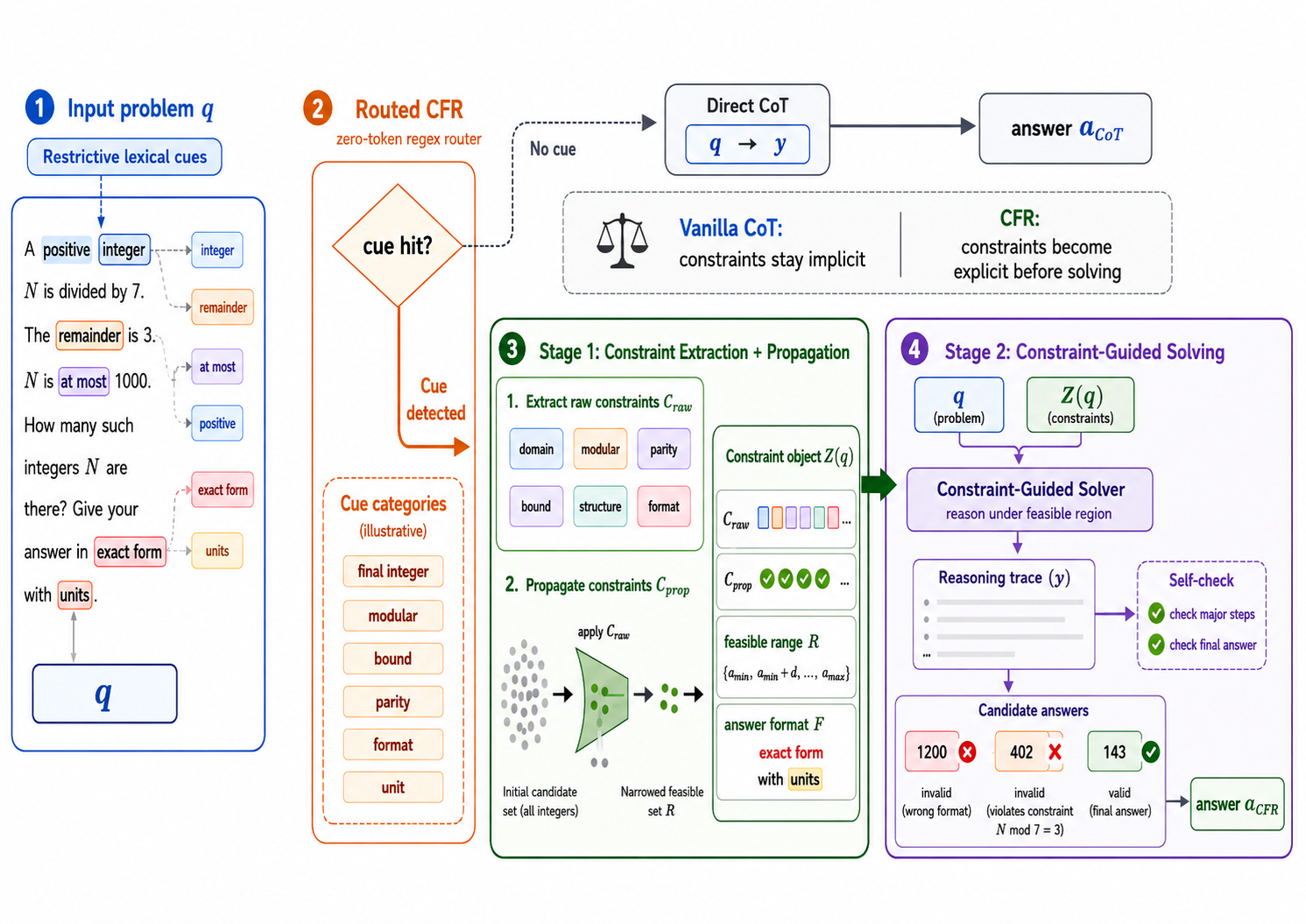}
\caption{\textbf{\rcfr{} framework overview.} Given an input problem, the regex router checks for restrictive lexical cues. If detected, Stage~1 extracts and cross-propagates answer-space constraints (domain, modular, parity, bound, format), then Stage~2 solves under the resulting feasible set with interleaved self-checks. Problems without cues bypass the pipeline via direct CoT.}
\label{fig:framework}
\end{figure*}

%==================================================================
\section{Related Work}
\label{sec:related}
%==================================================================

\subsection{Self-Refinement, Planning, and Verification}

Self-Refine~\citep{madaan2023selfrefine} iteratively improves outputs via self-feedback. Process reward models~\citep{lightman2023lets,wang2024mathshepherd} score individual reasoning steps to guide search. Outcome reward models~\citep{cobbe2021gsm8k} provide binary signals on final answers. ReAct~\citep{yao2023react} interleaves reasoning with actions for grounded decision-making, while Plan-and-Solve~\citep{wang2023plansolve} produces an explicit plan before solution generation. Learning-based alternatives further examine robust preference optimization, evolving sample interactions for multi-domain fine-tuning, noisy or verifier-independent reasoning rewards, and policy optimization through interactive guidance~\citep{liang2025ropo,liang2025boosting,cai2025reinforcement,cai2026vi,liu2026socraticpo}. These methods respectively emphasize planning, post-hoc refinement, verification, or policy learning. CFR is narrower: it constructs a problem-specific constraint summary before solving and carries that summary into the solving prompt. It therefore does not claim to replace planning, verification, or training-based optimization pipelines; Table~\ref{tab:prompting-main} empirically separates this design from their prompting counterparts.

\subsection{Test-Time Compute Scaling}

Recent work demonstrates that allocating more inference-time computation improves performance across diverse tasks. Best-of-$n$ sampling~\citep{brown2024large} and compute-optimal scaling~\citep{snell2024scaling} establish scaling laws for test-time compute. S1~\citep{muennighoff2025s1} shows that budget forcing (extending reasoning with ``wait'' tokens) improves accuracy. Unlike generic compute scaling, \cfr{} concentrates additional computation on constraint identification---a narrow task with compact output that produces high-value information guiding the larger solving budget.

%==================================================================
\section{Method}
\label{sec:method}
%==================================================================

\subsection{Problem Formulation}

Let $q$ denote a mathematical problem, $\mathcal{A}$ the answer space, $y$ a reasoning trace produced by the model, and $\pi(y)$ the function that extracts a final answer from $y$. Standard inference samples:
\begin{equation}
a^* = \pi(y^*), \quad y^* = \underset{y}{\arg\max}\; P_\theta(y \mid q)
\end{equation}
where $P_\theta$ is the language model. Many competition problems define a feasible answer region $\Phi(q) \subset \mathcal{A}$ through explicit constraints (e.g., ``integer in 000--999'' implies $|\Phi(q)| = 1000$ vs.\ $|\mathcal{A}| = \infty$). \cfr{} makes $\Phi(q)$ explicit:
\begin{equation}
a^*_{\text{CFR}} = \underset{a \in \Phi(q)}{\arg\max}\; P_\theta(a \mid q, \Phi(q))
\label{eq:cfr}
\end{equation}

The feasible-set notation is a conceptual description of the goal rather than a claim of formal constrained decoding. Because the Stage~1 summary is generated from $q$, it contributes no external evidence; its potential value is to reorganize problem information into an explicit prompt context that can change the model's subsequent reasoning behavior. We therefore treat the information-theoretic discussion in Appendix~\ref{app:math-framework} as intuition, not as a derived performance guarantee.

\subsection{Stage 1: Constraint Extraction and Prompted Summary}

Given problem $q$, Stage~1 prompts the model to extract structured constraints without solving, following a fixed taxonomy:

We distinguish \emph{direct answer-space constraints}, which directly restrict the final form, range, modularity, or option set, from \emph{problem-structure constraints}, such as dimensional consistency, geometric invariants, and branch conditions. The latter do not necessarily specify the final answer form, but can rule out invalid reasoning paths and thereby indirectly narrow admissible answers.

\begin{itemize}[nosep,leftmargin=*]
    \item \textbf{Domain}: integrality, positivity, ranges
    \item \textbf{Modular}: remainder conditions, divisibility, congruences
    \item \textbf{Parity}: even/odd requirements
    \item \textbf{Bound}: inequalities, extremal values
    \item \textbf{Monotonic}: increasing/decreasing relationships
    \item \textbf{Structure}: symmetry, counting targets, geometric invariants
    \item \textbf{Dimension}: unit consistency, coordinate ranges
    \item \textbf{Format}: encoded answers ($p+q$, remainder, coprime form)
\end{itemize}

After extraction, the model derives a prompted constraint summary. For example, if $C_1$: ``$n$ is divisible by 3'' and $C_2$: ``$100 \leq n \leq 200$,'' the summary may state that valid candidates are multiples of three between 102 and 198. The output takes the form of a structured JSON specification containing raw constraints, derived implications, the narrowest feasible range, and the required answer format.

\emph{Terminology.} ``Propagation'' here refers only to prompt-driven natural-language summarization and checking by the same frozen LLM, not to symbolic constraint propagation or a formal constraint-satisfaction procedure. We claim neither soundness nor completeness, and evaluate the resulting reliability directly in Section~\ref{sec:constraint-reliability}.

\paragraph{Token budget.} Stage~1 is allocated a concise budget so that constraint extraction remains compact relative to the final solving stage.

\subsection{Stage 2: Constraint-Guided Solving}

Stage~2 receives the original problem $q$ and the constraint specification from Stage~1. The solver must:
\begin{enumerate}[nosep]
    \item Reason step by step toward a solution.
    \item After each major step, check consistency against the constraints.
    \item If an intermediate result violates a constraint, stop and re-examine.
    \item Before stating the final answer, confirm all constraints are satisfied.
\end{enumerate}

This contrasts with self-correction approaches ($q \!\to\! y \!\to\! \text{critique}(y)$), which verify \emph{after} generation. \cfr{} instead front-loads constraint awareness ($q \!\to\! C(q) \!\to\! y$), providing multiple opportunities to catch violations \emph{during} reasoning.

\subsection{Routed CFR}
\label{sec:router}

Not all problems benefit from constraint extraction. We introduce a regex-based router that activates \cfr{} only when lexical cues indicate exploitable constraints.

The default router uses only the problem text; benchmark identity and benchmark-level answer conventions are not passed to either Stage~1 or Stage~2. The router consumes zero model tokens---it is a pure regex pass over the problem text. We separately audit the effect of the historical AIME convention override in Section~\ref{sec:router-controls}; that ablation clarifies a routing prior without supplying solver-side information.

\begin{algorithm}[t]
\caption{\rcfr{} Inference Protocol}
\label{alg:routed-cfr}
\small
\begin{algorithmic}[1]
\REQUIRE Problem $q$, model $P_\theta$, budget $B$
\STATE $\text{cues} \leftarrow \textsc{RegexRouter}(q)$
\IF{$|\text{cues}| > 0$}
    \STATE $C \leftarrow P_\theta(\text{extract\_constraints} \mid q)$ \hfill {\color{gray}$\triangleright$ Stage 1}
    \STATE $a \leftarrow P_\theta(\text{solve} \mid q, C)$ \hfill {\color{gray}$\triangleright$ Stage 2}
\ELSE
    \STATE $a \leftarrow P_\theta(\text{solve} \mid q)$ \hfill {\color{gray}$\triangleright$ Direct CoT}
\ENDIF
\RETURN $a$
\end{algorithmic}
\end{algorithm}

%==================================================================
\section{Experimental Setup}
\label{sec:setup}
%==================================================================

\subsection{Models}

We evaluate four models spanning a wide capability range:
\begin{itemize}[nosep,leftmargin=*]
    \item \textbf{DeepSeek-V4-Pro}~\citep{deepseekv4}: Frontier-class reasoning model.
    \item \textbf{Qwen3.5-35B-A3B}~\citep{qwen3.5}: Strong MoE model.
    \item \textbf{Qwen3.5-9B}~\citep{qwen3.5}: Mid-size dense model.
    \item \textbf{JustRL-1.5B}~\citep{he2025justrl}: Small RL-trained reasoning model.
\end{itemize}

\begin{table*}
  \centering
  \begin{tabular}{lccccc}
    \hline
    \textbf{Model} & \textbf{AIME24} & \textbf{AIME25} & \textbf{AIME26} & \textbf{CMIMC25} & \textbf{Avg.} \\
    \hline
    \multicolumn{6}{c}{\cellcolor{gray!15}\textbf{Baselines: Math LLMs}} \\
    \hline
    Qwen3-1.7B        &      40.2\%           &      46.7\%           &     38.1\%            &           20.0\%      &        36.3\%       \\
    Qwen3-4B-Thinking-2507        &        83.3\%         &          66.5\%                     &       70.3\%          &          57.5\%    & 69.4\% \\
    Qwen3-8B       &         73.6\%        &         71.5\%        &     69.2\%            &      52.2\%           &       66.6\%        \\
    \hline
    \multicolumn{6}{c}{\cellcolor{gray!15}\textbf{Baselines and Ours: JustRL-1.5B}} \\
    \hline
    JustRL-1.5B         & \textbf{47.5\%} & 37.5\%          & 40.0\%          & 17.5\%          & 35.6\%        \\
    Routed-CFR-JustRL-1.5B (Ours)    & \textbf{47.5\%}          & \textbf{45.8\%} & \textbf{42.5\%} & \textbf{18.1\%} & \textbf{38.5\%} \\
    \hline
    $\Delta$ \textit{vs.} JustRL-1.5B 
                        & 0.0\% 
                        & \textcolor{OliveGreen}{+8.3\%} 
                        & \textcolor{OliveGreen}{+2.5\%} 
                        & \textcolor{OliveGreen}{+0.6\%} 
                        & \textcolor{OliveGreen}{+2.9\%} \\
    \hline
    \multicolumn{6}{c}{\cellcolor{gray!15}\textbf{Baselines and Ours: Qwen3.5-35B-A3B}} \\
    \hline
    Qwen3.5-35B-A3B     & 87.5\%          & 83.3\%          & 87.5\%          & 30.6\%          & 72.2\%        \\
    Routed-CFR-Qwen3.5-35B-A3B (Ours) & \textbf{90.0\%} & \textbf{84.2\%} & \textbf{92.3\%} & \textbf{34.3\%} & \textbf{75.2\%} \\
    \hline
    $\Delta$ \textit{vs.} Qwen3.5-35B-A3B  
                        & \textcolor{OliveGreen}{+2.5\%}  
                        & \textcolor{OliveGreen}{+0.9\%}  
                        & \textcolor{OliveGreen}{+4.8\%}  
                        & \textcolor{OliveGreen}{+3.7\%}  
                        & \textcolor{OliveGreen}{+3.0\%}  \\
    \hline
    \multicolumn{6}{c}{\cellcolor{gray!15}\textbf{Baselines and Ours: Qwen3.5-9B}} \\
    \hline
    Qwen3.5-9B          & 77.5\%          & 61.7\%          & 65.0\%          & 20.0\%          & 56.1\%        \\
    Routed-CFR-Qwen3.5-9B (Ours)     & \textbf{81.7\%} & \textbf{70.8\%} & \textbf{78.3\%} & \textbf{22.5\%} & \textbf{63.3\%} \\
    \hline
    $\Delta$ \textit{vs.} Qwen3.5-9B      
                        & \textcolor{OliveGreen}{+4.2\%}  
                        & \textcolor{OliveGreen}{+9.1\%} 
                        & \textcolor{OliveGreen}{+13.3\%} 
                        & \textcolor{OliveGreen}{+2.5\%}  
                        & \textcolor{OliveGreen}{+7.2\%}  \\
    \hline
    \multicolumn{6}{c}{\cellcolor{gray!15}\textbf{Baselines and Ours: DeepSeek-V4-Pro}} \\
    \hline
    DeepSeek-V4-Pro    & 90.8\%          & 80.8\%          & 83.3\%          & 28.1\%          & 70.8\%        \\
    Routed-CFR-DeepSeek-V4-Pro (Ours) & \textbf{95.8\%} & \textbf{91.7\%} & \textbf{92.5\%} & \textbf{36.9\%} & \textbf{79.2\%} \\
    \hline
    $\Delta$ \textit{vs.} DeepSeek-V4-Pro  
                        & \textcolor{OliveGreen}{+5.0\%}  
                        & \textcolor{OliveGreen}{+10.9\%} 
                        & \textcolor{OliveGreen}{+9.2\%}  
                        & \textcolor{OliveGreen}{+8.8\%}  
                        & \textcolor{OliveGreen}{+8.5\%}  \\
    \hline
  \end{tabular}
  \caption{\label{tab:model-performance-main} 
    Convention-controlled performance comparison across AIME and CMIMC. Routed-CFR uses the text-only router; the AIME benchmark convention is disabled. All values are pass@1 averaged over four independent runs. $\Delta$ denotes the gain over the matched direct-CoT backbone.
  }
\end{table*}

\begin{table*}[!ht]
  \centering
  \setlength{\tabcolsep}{6pt}
  \begin{tabular}{l *{3}{cc}}
    \hline
    \textbf{Model} & \multicolumn{2}{c}{\textbf{BRUMO25}} & \multicolumn{2}{c}{\textbf{AIMO\_AMC}} & \multicolumn{2}{c}{\textbf{CMIMC25}} \\
    \cmidrule(lr){2-3} \cmidrule(lr){4-5} \cmidrule(lr){6-7}
    & \textbf{Acc.} & \textbf{Token} & \textbf{Acc.} & \textbf{Token} & \textbf{Acc.} & \textbf{Token} \\
    \hline
    DeepSeek-V4-Pro & 56.7\% & 1.00x & 97.5\% & 1.00x & 28.1\% & 1.00x \\
    \hline
    Routed-CFR-DeepSeek-V4-Pro (Ours) & \textbf{60.0\%} & \textbf{1.58x} & \textbf{97.5\%} &  \textbf{2.08x} & \textbf{36.9\%} & \textbf{1.64x} \\
    CFR-DeepSeek-V4-Pro (always-on) & 57.8\% & 1.71x & 96.1\% & 2.45x & 34.2\% &  1.94x\\
    \hline
  \end{tabular}
  \caption{\label{tab:model-performance-sub} 
    Routing ablation on DeepSeek-V4-Pro. Routed-CFR uses the text-only router; CFR (always-on) runs both stages for every problem. Token values are relative to direct CoT on the same benchmark.
  }
\end{table*}

\subsection{Evaluation Scope}

The empirical section uses complementary evidence rather than a single headline table. Table~\ref{tab:model-performance-main} is the convention-controlled multi-backbone comparison, Table~\ref{tab:prompting-main} combines matched prompting baselines with total-token cost, and Table~\ref{tab:model-performance-sub} is the routing ablation. Detailed paired statistical analyses, benchmark descriptions, router audits, and error analyses are provided in the appendix. Table~\ref{tab:model-performance-additional} extends BRUMO25 and AIMO\_AMC to all reported backbones.

\paragraph{Methods compared.} Direct CoT is the primary baseline. Strong prompting alternatives are compared under the same protocol in Table~\ref{tab:prompting-main}. Routed-CFR uses text-only routing, while CFR (always-on) is used only for the routing ablation.

%==================================================================
\section{Results}
\label{sec:results}
%==================================================================

\subsection{Main Results}

We begin with the convention-controlled multi-backbone comparison, then narrow the discussion to prompting controls and cost. The appendix describes the benchmark characteristics: AIME has a tight integer answer convention, while BRUMO, AIMO\_AMC, and CMIMC retain explicit local constraints but vary in format and difficulty.

Table~\ref{tab:model-performance-main} provides the central result. With the AIME convention removed from routing, Routed-CFR improves the AIME/CMIMC average for each evaluated backbone. The gains are smallest for the low-capacity model and largest for DeepSeek-V4-Pro and Qwen3.5-9B, consistent with the reliability analysis in Section~\ref{sec:constraint-reliability}. Figure~\ref{fig:gains-by-model} visualizes the model--benchmark interaction, and Table~\ref{tab:model-performance-additional} extends the comparison to BRUMO25 and AIMO\_AMC.

Taken together, these performance tables support a conditional claim rather than a universal one: CFR can help when the problem contains recoverable constraints and the backbone can use the Stage~1 summary. Points near ceiling accuracy naturally have little room to move, while lower-capacity models can be harmed by incomplete or invalid extraction.

\subsection{Matched Prompting Baselines}

Table~\ref{tab:prompting-main} compares CFR with stronger training-free prompting alternatives on DeepSeek-V4-Pro under the same pass@1 avg@4 protocol. These controls separate constraint-first guidance from a final-answer reminder, simply listing constraints, longer free-form reasoning, planning, format-only extraction, and post-hoc refinement. Text-only Routed-CFR is best on AIME24 and AIME25, ties the best AIME26 score, and remains competitive on the near-ceiling AIMO\_AMC setting.

\begin{table*}[t]
\centering
\scriptsize
\setlength{\tabcolsep}{3pt}
\resizebox{\textwidth}{!}{%
\begin{tabular}{lrrrrrrrr}
\toprule
\textbf{Method} & \multicolumn{4}{c}{\textbf{Accuracy (\%)}} & \multicolumn{4}{c}{\textbf{Mean tokens per sample}} \\
\cmidrule(lr){2-5}\cmidrule(lr){6-9}
& \textbf{AIME24} & \textbf{AIME25} & \textbf{AIME26} & \textbf{AIMO\_AMC} & \textbf{Input} & \textbf{Output} & \textbf{Total} & \textbf{vs. CFR} \\
\midrule
Direct CoT & 90.8 & 80.8 & 83.3 & 97.5 & 305 & 13,956 & 14,261 & 0.58$\times$ \\
CoT + final-answer constraint reminder & 87.5 & 88.3 & 89.2 & 96.9 & --- & --- & --- & --- \\
List-constraints-first & 91.7 & 89.2 & 89.2 & \textbf{98.5} & --- & --- & --- & --- \\
Extended CoT (token-matched) & 91.7 & 89.2 & 91.7 & 97.4 & 321 & 21,294 & 21,615 & 0.88$\times$ \\
Plan-and-Solve & 90.0 & 90.8 & 91.7 & 98.0 & --- & --- & --- & --- \\
Format-only extraction + solve & 86.7 & 87.5 & \textbf{92.5} & \textbf{98.5} & 679 & 21,428 & 22,107 & 0.90$\times$ \\
Self-Refine & 89.2 & 85.8 & 87.5 & 95.4 & 3,801 & 33,044 & 36,845 & 1.50$\times$ \\
CFR (always-on)$^{\ddagger}$ & 88.9 & 88.9 & 91.1 & 98.0 & 1,927 & 25,338 & 27,265 & 1.11$\times$ \\
\textbf{Text-only Routed-CFR (ours)} & \textbf{95.8} & \textbf{91.7} & \textbf{92.5} & 97.5 & 1,736 & 22,336 & 24,072 & 0.98$\times$ \\
\bottomrule
\end{tabular}}
\caption{Matched prompting and accuracy--cost comparison on DeepSeek-V4-Pro. Accuracy values are pass@1 averaged over four independent runs. Token columns are mean per-sample profiling measurements; ``---'' denotes methods for which cost profiling was not available. $^{\ddagger}$The always-on row is included only to quantify the routing cost trade-off.}
\label{tab:prompting-main}
\label{tab:cost-accounting}
\end{table*}

\subsection{Convention Control and Paired Statistics}
\label{sec:paired-statistics}

To address the small size of the annual AIME sets, we conduct problem-level paired analyses after disabling the AIME-specific routing convention. For every problem, we average four independent pass@1 outcomes, compute the within-problem difference, construct paired bootstrap confidence intervals, and apply a two-sided paired randomization test. The full DeepSeek-V4-Pro and Qwen3.5-9B results appear in Appendix~\ref{app:paired-statistics}. The pooled gains remain significant after Holm correction for both models; individual annual sets have wider intervals and are therefore interpreted as supporting evidence rather than independent confirmation of a universal effect.

\subsection{Ablation Results}

Table~\ref{tab:model-performance-sub} isolates the routing decision for DeepSeek-V4-Pro. The token multipliers explain why routing is not merely a convenience: Routed-CFR uses lower relative token cost than the always-on variant on every reported benchmark ($1.58\times$ vs. $1.71\times$ on BRUMO25, $2.08\times$ vs. $2.45\times$ on AIMO\_AMC, and $1.64\times$ vs. $1.94\times$ on CMIMC25). Thus, routing can improve the accuracy--cost trade-off by applying the full protocol only when the problem text suggests useful constraints.

\subsection{Interpretation}

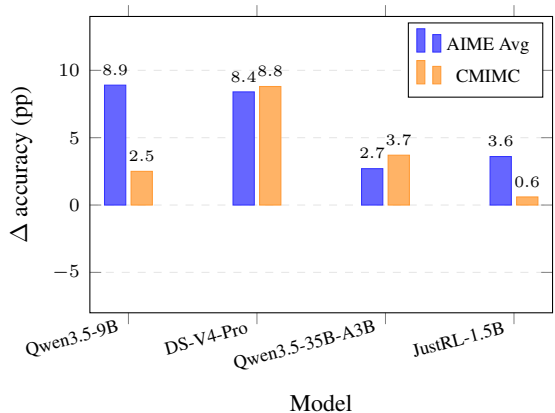
\begin{figure}[t]
\centering
\begin{tikzpicture}
\begin{axis}[
    width=\columnwidth,
    height=5.5cm,
    ybar,
    bar width=8pt,
    xlabel={Model},
    ylabel={$\Delta$ accuracy (pp)},
    ymin=-8, ymax=14,
    xtick={1,2,3,4},
    xticklabels={Qwen3.5-9B, DS-V4-Pro, Qwen3.5-35B-A3B, JustRL-1.5B},
    x tick label style={font=\scriptsize, rotate=15, anchor=east},
    every axis label/.style={font=\small},
    tick label style={font=\scriptsize},
    legend style={font=\scriptsize, at={(0.98,0.98)}, anchor=north east},
    grid=major,
    grid style={dashed, gray!20},
    ymajorgrids=true,
    xmajorgrids=false,
    nodes near coords,
    every node near coord/.append style={font=\tiny},
]
\addplot[fill=blue!60, draw=blue!80] coordinates {(1,8.9) (2,8.4) (3,2.7) (4,3.6)};
\addplot[fill=orange!60, draw=orange!80] coordinates {(1,2.5) (2,8.8) (3,3.7) (4,0.6)};
\legend{AIME Avg, CMIMC}
\end{axis}
\end{tikzpicture}
\caption{\cfr{} gain ($\Delta$ accuracy) by model, computed directly from Table~\ref{tab:model-performance-main}. AIME values are the mean of the three annual AIME gains; CMIMC values are the corresponding CMIMC25 gains.}
\label{fig:gains-by-model}
\end{figure}

The remaining figures and tables explain the pattern behind the headline numbers. Figure~\ref{fig:gains-by-model} separates gains by model, while the appendix provides benchmark characteristics and a detailed applicability plot. Together with the constraint audit in Section~\ref{sec:constraint-reliability}, the evidence supports a limited interpretation: constraints are helpful when they are explicit enough to recover, restrictive enough to reject plausible wrong answers, and paired with a backbone capable of using them.

The bar plot in Figure~\ref{fig:gains-by-model} makes the model--benchmark interaction easier to read than the raw table. Qwen3.5-9B and DeepSeek-V4-Pro show the largest controlled AIME gains, whereas the other backbones have smaller but positive average gains. This is an empirical pattern, not a derived scaling law: Section~\ref{sec:constraint-reliability} shows that Stage~1 quality is a plausible moderator of the observed effects.

%==================================================================
\section{Analysis}
\label{sec:analysis}
%==================================================================

The results above show where CFR helps; the analysis below examines when this conclusion is credible. We audit the router, the scope of text-only routing beyond AIME, Stage~1 reliability, total token cost, and the error categories associated with successful interventions.

\subsection{Router Behavior}
\label{sec:router-controls}

The AIME answer convention can be a routing prior, but it must not be confused with solver-side information. The convention was never inserted into Stage~1, Stage~2, or direct-CoT prompts. More importantly, the controlled results in this paper disable it entirely: the text-only router still activates for 83 of 90 AIME problems. The detailed routing control and paired tests are reported in Appendix~\ref{app:router-and-generalization}.

Table~\ref{tab:router-rates} extends the audit to benchmarks with heterogeneous answer requirements. The non-trivial activation rates on GSM8K, MATH-500, HMMT, and OlympiadBench show that the router is not restricted to a fixed 000--999 answer convention; they do not, by themselves, establish that every routed problem benefits from CFR.

\begin{table}[t]
\centering
\small
\setlength{\tabcolsep}{4pt}
\begin{tabular}{lrr}
\toprule
\textbf{Benchmark} & \textbf{Routed / total} & \textbf{Rate} \\
\midrule
GSM8K & 834 / 1,319 & 63.23\% \\
MATH-500 & 281 / 500 & 56.20\% \\
AMC23 & 32 / 40 & 80.00\% \\
HMMT 2025 & 21 / 30 & 70.00\% \\
OlympiadBench & 417 / 674 & 61.87\% \\
\bottomrule
\end{tabular}
\caption{External audit of the text-only router.}
\label{tab:router-rates}
\end{table}

\subsection{Generalization beyond Unified Answer Conventions}

The AIME convention is especially compatible with answer-space constraints, so AIME alone cannot establish generalization. We therefore evaluate on OlympiadBench with all benchmark-specific conventions disabled. Its heterogeneous problem sources and answer formats provide a complementary setting in which routing depends only on problem text: Text-only Routed-CFR reaches 91.14\% versus 90.66\% for direct CoT, with a 61.87\% routing rate. The detailed result appears in Appendix~\ref{app:router-and-generalization}; Table~\ref{tab:model-performance-additional} provides further evidence on CMIMC25, BRUMO25, and AIMO\_AMC. These results do not claim effectiveness on proof-oriented tasks; rather, they show that the mechanism can apply when local domain, modular, range, unit, encoding, or structural restrictions are recoverable without a shared benchmark convention.

\begin{table*}[t]
\centering
\footnotesize
\setlength{\tabcolsep}{4pt}
\renewcommand{\arraystretch}{1.05}
\caption{A representative positive case where CFR corrects CoT on DeepSeek-V4-Pro.}
\label{tab:cfr_positive_case_geometry}
\begin{tabularx}{\textwidth}{@{}>{\raggedright\arraybackslash}X>{\raggedright\arraybackslash}X@{}}
\toprule
\multicolumn{2}{p{\dimexpr\textwidth-2\tabcolsep\relax}}{
\textbf{Example: Ratio-and-Reflection Geometry (AIME 2025)}
\hfill \textcolor{teal}{31.1\% token reduction}} \\
\multicolumn{2}{p{\dimexpr\textwidth-2\tabcolsep\relax}}{
\textbf{Problem.} In $\triangle ABC$, points $D,E$ lie on $\overline{AB}$ and $F,G$ lie on $\overline{AC}$.
Given $AD=4$, $DE=16$, $EB=8$, $AF=13$, $FG=52$, $GC=26$, let $M$ be the reflection of $D$ through $F$ and $N$ the reflection of $G$ through $E$.
If $[DEGF]=288$, find $[AFNBCEM]$.} \\
\multicolumn{2}{p{\dimexpr\textwidth-2\tabcolsep\relax}}{
\textbf{Ground Truth:} $588$ \qquad \textbf{CoT:} 33,065 tokens \qquad \textbf{CFR:} 22,791 tokens} \\
\midrule
\textbf{DeepSeek-V4-Pro CoT} & \textbf{Routed-CFR-DeepSeek-V4-Pro} \\
\midrule
\textcolor{red}{CoT notices that $D,F$ and $E,G$ divide the two sides in the same ratios, but keeps the solution as a free-form geometric narrative. It never turns $[DEGF]=288$ into the decisive constraint on $\sin A$, and the reflected points $M,N$ are not explicitly encoded. The heptagon area therefore drifts away from the constrained construction. \textbf{Answer:} $\boxed{1}$ $\times$}
&
\textcolor{teal}{CFR makes the constraints explicit before solving:
\(\frac{AD}{AB}=\frac{AF}{AC}=\frac17,\ \frac{AE}{AB}=\frac{AG}{AC}=\frac57\);
\([DEGF]=624\sin A=288\Rightarrow \sin A=\frac6{13}\);
and \(M=2F-D,\ N=2E-G\). Shoelace then gives
\([AFNBCEM]=1274\sin A=1274\cdot\frac6{13}=588\).
\textbf{Answer:} $\boxed{588}$ \checkmark}
\\
\bottomrule
\end{tabularx}
\end{table*}

\begin{table*}[t]
\centering
\footnotesize
\setlength{\tabcolsep}{4pt}
\renewcommand{\arraystretch}{1.05}
\caption{A representative positive case where CFR corrects CoT on Qwen3.5-9B.}
\label{tab:qwen_positive_case_geometry}
\begin{tabularx}{\textwidth}{@{}>{\raggedright\arraybackslash}X>{\raggedright\arraybackslash}X@{}}
\toprule
\multicolumn{2}{p{\dimexpr\textwidth-2\tabcolsep\relax}}{
\textbf{Example: Cyclic-Constraint Geometry (Qwen3.5-9B, AIME 2024)}
\hfill \textcolor{teal}{35.3\% token reduction}} \\
\multicolumn{2}{p{\dimexpr\textwidth-2\tabcolsep\relax}}{
\textbf{Problem.} Rectangles $ABCD$ and $EFGH$ are drawn such that $D,E,C,F$ are collinear. Also, $A,D,H,G$ all lie on a circle.
If $BC=16$, $AB=107$, $FG=17$, and $EF=184$, find $CE$.} \\
\multicolumn{2}{p{\dimexpr\textwidth-2\tabcolsep\relax}}{
\textbf{Ground Truth:} $104$ \qquad \textbf{CoT:} 33,015 tokens \qquad \textbf{CFR:} 21,370 tokens} \\
\midrule
\textbf{Qwen3.5-9B CoT} & \textbf{Routed-CFR-Qwen3.5-9B} \\
\midrule
\textcolor{red}{CoT sets up coordinates but does not enforce the relative vertical orientation of the two rectangles. It allows the second rectangle to remain on the same side of the baseline without checking concyclicity, so the circle constraint is applied to the wrong configuration. \textbf{Answer:} $\boxed{64}$ $\times$}
&
\textcolor{teal}{CFR fixes the feasible orientation:
\(D=(0,0), C=(107,0), A=(0,16)\), while concyclicity forces
\(E=(x,0), F=(x+184,0), H=(x,-17), G=(x+184,-17)\).
The circle through \(A,D\) has center on \(y=8\), and chord \(HG\) gives center \(x\)-coordinate \(x+92\), so
\((x+92)^2+8^2=92^2+25^2\). Thus \(x=3\) and \(CE=107-3=104\).
\textbf{Answer:} $\boxed{104}$ \checkmark}
\\
\bottomrule
\end{tabularx}
\end{table*}

\subsection{Constraint Category Analysis}

Next, we ask which extracted constraints are most responsible for the AIME gains. The appendix reports category frequency and the associated gain. Modular constraints show the largest descriptive marginal benefit (+8.7 pp), even though domain constraints occur in every AIME problem. This pattern suggests that frequency alone does not determine usefulness: modular and encoded-format constraints often require an explicit final transformation that benefits from being kept active throughout the solution.

\subsection{Stage~1 Constraint Reliability}
\label{sec:constraint-reliability}

The framework is training-free and can be applied to different backbones without parameter updates, but it is not performance-agnostic: useful gains depend on Stage~1 producing a sufficiently complete and valid constraint summary. We therefore manually compare extracted constraints with those entailed by the original problem across models of different capacities. Table~\ref{tab:extraction-reliability} distinguishes the number of initially extracted constraints, the number carried into Stage~2, the validity of carried constraints, and hallucination rate. The smallest model extracts fewer constraints and has much lower propagated validity; this is a more direct explanation of its weaker gains than a claim that CFR works uniformly across scales.

\begin{table}[t]
\centering
\scriptsize
\setlength{\tabcolsep}{3.5pt}
\begin{tabular}{lrrrr}
\toprule
\textbf{Model} & \textbf{Raw} & \textbf{Propagated} & \textbf{Validity} & \textbf{Halluc.} \\
\midrule
DeepSeek-V4-Pro & 7.74 & 5.12 & 97.5\% & 15.6\% \\
Qwen3.5-9B & 4.88 & 3.34 & 93.3\% & 13.9\% \\
Qwen3.5-35B-A3B & 5.00 & 3.64 & 91.8\% & 17.8\% \\
JustRL-1.5B & 2.86 & 2.19 & 73.9\% & 17.5\% \\
\bottomrule
\end{tabular}
\caption{Manual Stage~1 audit. Raw and propagated columns are mean counts per problem; validity and hallucination are assessed against constraints entailed by the problem statement.}
\label{tab:extraction-reliability}
\end{table}

\subsection{Token Overhead}

The next question is cost. Output tokens alone omit the additional Stage~1 and Stage~2 prompt context, so the merged performance--cost comparison in Table~\ref{tab:prompting-main} reports input, output, and total tokens. The two-stage protocol costs more than direct CoT, while routing reduces the overhead relative to always-on CFR.

The overhead is not uniform across backbones. DeepSeek-V4-Pro has concise baseline traces, so the additional constraint stage is proportionally large; Qwen3.5-35B-A3B already produces longer baseline reasoning, so the same kind of constraint prefix is a smaller relative increase. The practical conclusion is deliberately limited: routing can reduce unnecessary cost, but it does not make CFR free and should be considered only where its targeted benefit justifies the added tokens.

\subsection{Capability Interaction: An Empirical Interpretation}
\label{sec:diminishing}

The appendix applicability plot suggests that benefits depend jointly on problem difficulty, constraint strength, and Stage~1 reliability. Near-ceiling settings leave little room for improvement, whereas a weak model may not produce a reliable enough summary to help the solver. This is an empirical interpretation of the observed pattern, not a theoretical derivation or a claim that medium-capability models necessarily benefit most. A formal account would require an explicit interaction analysis and a model of extraction error, which are beyond the present study.

\subsection{Error Analysis}
\label{sec:error-analysis}

Finally, the appendix error analysis examines what kinds of mistakes CFR actually fixes. We manually analyze the AIME 2024--2026 cases where CFR flips Qwen3.5-9B from an incorrect to a correct outcome.

The error distribution confirms the central mechanism. Most fixes are not cases where CFR discovers a wholly new solution strategy; they are cases where the baseline reaches a plausible mathematical object but violates the required answer format, misses a modular reduction, or fails to encode the final response correctly. The smaller arithmetic category shows a secondary benefit: once the constraints are explicit, intermediate checks can catch calculations that would otherwise slip through to the final answer.

%==================================================================
\section{Case Studies}
\label{sec:cases}

The aggregate analyses above identify the mechanism statistically; the two case studies show it operationally. Both examples involve ordinary-looking geometry problems where the baseline begins with a plausible setup but loses a decisive constraint. CFR changes the trajectory by making the hidden feasibility conditions explicit before the final derivation begins.

Table~\ref{tab:cfr_positive_case_geometry} shows the first failure mode: the baseline notices the proportional structure but does not convert the area and reflection conditions into binding equations. CFR instead turns the ratios, the area constraint, and the reflected-point definitions into a compact coordinate system, making the final area computation well determined. Table~\ref{tab:qwen_positive_case_geometry} shows a related geometric-branch failure. The baseline sets coordinates but keeps the wrong orientation feasible; CFR uses concyclicity to fix the orientation before solving. Together, the cases illustrate why the quantitative gains concentrate on constraint-heavy benchmarks: the method does not merely ask for longer reasoning, but changes which facts remain active while the model reasons.

%==================================================================

%==================================================================
\section{Conclusion}
\label{sec:conclusion}
%==================================================================

We introduced \cfr{}, a training-free protocol that extracts a problem-specific constraint summary before solving, and \rcfr{}, a text-only router that activates it selectively. Under convention-controlled routing, Routed-CFR improves the reported AIME/CMIMC averages by $+2.9$ to $+8.5$ percentage points across the four evaluated backbones. Matched prompting baselines, paired bootstrap intervals and randomization tests, an external router audit, a Stage~1 quality analysis, and complete token accounting strengthen the interpretation of these gains. The evidence does not support a general reasoning guarantee: CFR incurs added cost, is sensitive to extraction reliability, and is most appropriate when explicit answer-space or problem-structure constraints can guide the reasoning trace. Its principal effect is the prevention of answer-format, modular, and range violations rather than generic reasoning improvement.

\clearpage

\section*{Limitations}

The empirical claims are limited to the reported numerical-answer benchmarks and the documented prompting protocols. They do not establish a universal scaling law, formal constraint-solving guarantee, or improvement on open-ended proof tasks; the latter remain particularly challenging for current language models. CFR also depends on valid and sufficiently complete Stage~1 extraction. If the summary is incomplete or incorrect, it can bias Stage~2 toward an invalid feasible region, as suggested by the lower propagated validity for JustRL-1.5B. Finally, the router audit measures activation, not calibrated probability of benefit, and the two-stage protocol increases both input and output token use. Future work should evaluate learned or calibrated routers, annotate extraction completeness at larger scale, and test proof-oriented and open-answer settings.

%==================================================================
% REFERENCES
%==================================================================
\bibliography{custom}

%==================================================================
% APPENDIX
%==================================================================
\clearpage
\appendix

\section{Supplementary Empirical Evaluation}
\label{app:empirical-evaluation}

\subsection{Convention-Controlled Paired Statistics}
\label{app:paired-statistics}

We report the complete problem-level paired analyses underlying the convention-controlled results in Section~\ref{sec:paired-statistics}.

\begin{table*}[t]
\centering
\scriptsize
\resizebox{\textwidth}{!}{%
\begin{tabular}{lrrrrlrr}
\toprule
\textbf{Benchmark} & \textbf{Direct CoT} & \textbf{Text-only Routed-CFR} & \textbf{Gain} & \textbf{Improved / Tied / Degraded} & \textbf{95\% paired CI} & \textbf{Paired $p$} & \textbf{Holm $p$} \\
\midrule
AIME24 & 90.8 & 95.8 & 5.0 & 3 / 27 / 0 & [0.8, 15.0] & 0.25 & 0.25 \\
AIME25 & 80.8 & 91.7 & 10.8 & 8 / 22 / 0 & [5.0, 22.5] & 0.0078 & \textbf{0.0313} \\
AIME26 & 83.3 & 92.5 & 9.2 & 6 / 23 / 1 & [2.5, 19.2] & 0.0469 & 0.0938 \\
AIME24--26 pooled & 85.0 & 93.3 & 8.3 & 17 / 72 / 1 & [4.4, 12.8] & 0.000069 & \textbf{0.00034} \\
\bottomrule
\end{tabular}}
\caption{Problem-level paired analysis for DeepSeek-V4-Pro after removing the AIME routing convention. Each problem contributes its mean pass@1 accuracy over four runs.}
\label{tab:paired-ds}
\end{table*}

\begin{table*}[t]
\centering
\scriptsize
\resizebox{\textwidth}{!}{%
\begin{tabular}{lrrrrlrr}
\toprule
\textbf{Benchmark} & \textbf{Direct CoT} & \textbf{Text-only Routed-CFR} & \textbf{Gain} & \textbf{Improved / Tied / Degraded} & \textbf{95\% paired CI} & \textbf{Paired $p$} & \textbf{Holm $p$} \\
\midrule
AIME24 & 77.5 & 81.7 & 4.2 & 6 / 20 / 4 & [-3.3, 14.2] & 0.4590 & 0.4590 \\
AIME25 & 61.7 & 70.8 & 9.2 & 8 / 19 / 3 & [1.7, 20.8] & 0.0859 & 0.1719 \\
AIME26 & 65.0 & 78.3 & 13.3 & 8 / 20 / 2 & [5.0, 25.8] & 0.0234 & 0.0703 \\
AIME24--26 pooled & 68.1 & 76.9 & 8.9 & 22 / 59 / 9 & [3.6, 14.4] & 0.00198 & \textbf{0.00793} \\
\bottomrule
\end{tabular}}
\caption{Problem-level paired analysis for Qwen3.5-9B after removing the AIME routing convention. Each problem contributes its mean pass@1 accuracy over four runs.}
\label{tab:paired-qwen}
\end{table*}

\clearpage
\subsection{Benchmark Scope and Router Behavior}
\label{app:router-and-generalization}

The following tables document the answer-space restrictions across benchmarks and the effect of retaining or removing the AIME-specific routing convention.

\begin{table}[t]
\centering
\small
\setlength{\tabcolsep}{4pt}
\begin{tabular}{lcll}
\toprule
\textbf{Benchmark} & \textbf{Problems} & \textbf{Answer format} & \textbf{Constraint strength} \\
\midrule
AIME24 & 30 & Integer in 000--999 & Tight \\
AIME25 & 30 & Integer in 000--999 & Tight \\
AIME26 & 30 & Integer in 000--999 & Tight \\
AMC2023 & 40 & Multiple choice & Medium \\
BRUMO25 & 30 & Integer & Medium \\
CMIMC25 & 40 & Integer & Medium \\
OlympiadBench & 674 & Heterogeneous & Weak \\
\bottomrule
\end{tabular}
\caption{Benchmark characteristics. ``Constraint strength'' summarizes how tightly the answer format restricts the feasible answer space.}
\label{tab:benchmarks}
\end{table}

\begin{table}[t]
\centering
\small
\begin{tabular}{lrrr}
\toprule
\textbf{Router setting} & \textbf{AIME24} & \textbf{AIME25} & \textbf{AIME26} \\
\midrule
Text-only router & 28 / 30 & 28 / 30 & 27 / 30 \\
Text-only router + AIME convention & 30 / 30 & 30 / 30 & 30 / 30 \\
\bottomrule
\end{tabular}
\caption{AIME router convention control. The primary results use the text-only router without the AIME-specific convention.}
\label{tab:router-convention}
\end{table}

\clearpage
\subsection{Generalization and Applicability}
\label{app:applicability}

We next report the convention-free OlympiadBench result and summarize the benchmark-level applicability pattern for DeepSeek-V4-Pro.

\begin{table}[t]
\centering
\small
\begin{tabular}{lrrrr}
\toprule
\textbf{Model} & \textbf{Direct CoT} & \textbf{Text-only Routed-CFR} & \textbf{Gain} & \textbf{Routing rate} \\
\midrule
DeepSeek-V4-Pro & 90.66 & 91.14 & +0.48 & 417 / 674 (61.87\%) \\
\bottomrule
\end{tabular}
\caption{OlympiadBench generalization result (pass@1, \%). All benchmark-specific routing conventions are disabled.}
\label{tab:olympiadbench}
\end{table}

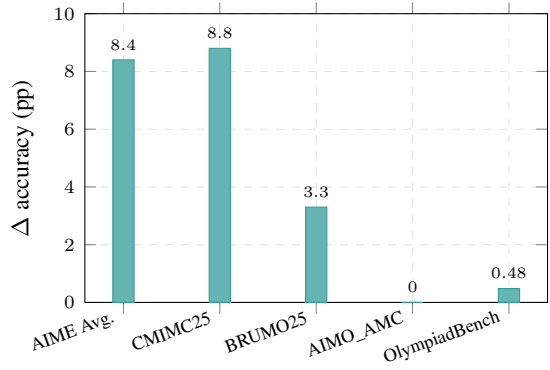
\begin{figure}[t]
\centering
\begin{tikzpicture}
\begin{axis}[
    width=\columnwidth,
    height=5.4cm,
    ybar,
    bar width=8pt,
    ylabel={$\Delta$ accuracy (pp)},
    ymin=0, ymax=10,
    xtick={1,2,3,4,5},
    xticklabels={AIME Avg., CMIMC25, BRUMO25, AIMO\_AMC, OlympiadBench},
    x tick label style={font=\scriptsize, rotate=20, anchor=east},
    tick label style={font=\scriptsize},
    every axis label/.style={font=\small},
    nodes near coords,
    every node near coord/.append style={font=\tiny},
    grid=major,
    grid style={dashed,gray!20},
]
\addplot[fill=teal!60,draw=teal!80] coordinates {(1,8.4) (2,8.8) (3,3.3) (4,0.0) (5,0.48)};
\end{axis}
\end{tikzpicture}
\caption{Applicability across benchmarks for DeepSeek-V4-Pro. Values are Text-only Routed-CFR gains over matched direct CoT; AIME is averaged across the three annual sets.}
\label{fig:applicability}
\end{figure}

\clearpage
\section{Supplementary Constraint-Level Analyses}
\label{app:constraint-and-error}

\subsection{Constraint Categories}

The table and figure below provide complementary tabular and visual summaries of the constraint types associated with CFR gains.

\begin{table}[t]
\centering
\small
\begin{tabular}{lrr}
\toprule
\textbf{Category} & \textbf{Frequency} & \textbf{$\Delta$ when present} \\
\midrule
Domain (integer) & 100\% & +5.2 pp \\
Modular (mod $n$) & 63\% & +8.7 pp \\
Format (encoded) & 41\% & +7.3 pp \\
Bound (range) & 38\% & +4.1 pp \\
Parity & 22\% & +3.8 pp \\
Structure & 18\% & +2.4 pp \\
\bottomrule
\end{tabular}
\caption{Constraint category frequency on AIME 2024--2026 and associated CFR gains for Qwen3.5-9B.}
\label{tab:constraint-categories}
\end{table}

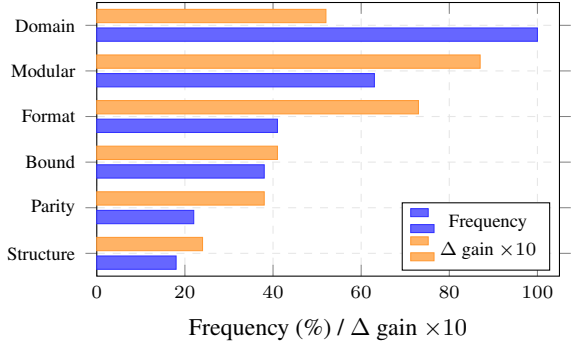
\begin{figure}[t]
\centering
\begin{tikzpicture}
\begin{axis}[
    width=\columnwidth,
    height=5.2cm,
    xbar,
    bar width=5pt,
    xmin=0, xmax=105,
    xlabel={Frequency (\%) / $\Delta$ gain $\times 10$},
    symbolic y coords={Domain,Modular,Format,Bound,Parity,Structure},
    ytick=data,
    y dir=reverse,
    tick label style={font=\scriptsize},
    every axis label/.style={font=\small},
    legend style={font=\scriptsize, at={(0.98,0.02)}, anchor=south east},
    grid=major,
    grid style={dashed,gray!20},
]
\addplot[fill=blue!60,draw=blue!80] coordinates {(100,Domain) (63,Modular) (41,Format) (38,Bound) (22,Parity) (18,Structure)};
\addplot[fill=orange!60,draw=orange!80] coordinates {(52,Domain) (87,Modular) (73,Format) (41,Bound) (38,Parity) (24,Structure)};
\legend{Frequency, $\Delta$ gain $\times 10$}
\end{axis}
\end{tikzpicture}
\caption{Constraint category distribution on AIME 2024--2026. The left bars show frequency of occurrence; the right bars show the associated CFR gain scaled by $10$ for visibility.}
\label{fig:constraint-categories}
\end{figure}

\clearpage
\subsection{Error Taxonomy}

\begin{table}[t]
\centering
\small
\begin{tabular}{lcc}
\toprule
\textbf{Error type fixed} & \textbf{Count} & \textbf{\%} \\
\midrule
Forgot modular reduction & 9 & 30\% \\
Non-integer answer & 6 & 20\% \\
Wrong encoded format ($p+q$) & 5 & 17\% \\
Range violation & 4 & 13\% \\
Arithmetic error caught by check & 3 & 10\% \\
Other & 3 & 10\% \\
\bottomrule
\end{tabular}
\caption{Types of errors fixed by \cfr{}. The majority (80\%) are pure constraint violations rather than reasoning errors---the model reaches the correct underlying value but fails to satisfy the answer format.}
\label{tab:error-analysis}
\end{table}

\clearpage
\section{Complete Prompt Templates}
\label{app:prompts}

\subsection{Constraint Extraction Prompt (Stage 1)}

\begin{tcolorbox}[colback=gray!5!white,colframe=gray!50!black,title=Stage 1 Prompt,fonttitle=\small\bfseries,fontupper=\scriptsize]
\texttt{You are a mathematical constraint analyst. Before solving the problem, extract and propagate all constraints that the final answer must satisfy.}

\texttt{Do not solve the problem. Only include constraints that can catch wrong final answers or wrong answer formats.}

\texttt{For each constraint, cross-check it against others to narrow the possible range. For example, if one constraint says ``n is divisible by 3'' and another says ``$100 \leq n \leq 200$,'' the propagated result is ``n is a multiple of 3 between 102 and 198 inclusive.''}

\texttt{Constraint types to look for:}\\
\texttt{- domain: ranges, sign, integrality, positivity}\\
\texttt{- modular: congruence, remainder, divisibility}\\
\texttt{- parity: even/odd}\\
\texttt{- bound: inequalities, extremal bounds}\\
\texttt{- monotonic: increasing/decreasing relationships}\\
\texttt{- structure: symmetry, combinatorial counts}\\
\texttt{- dimension: unit consistency, coordinate ranges}\\
\texttt{- format: exact form, encoded answer convention}

\texttt{Return JSON: \{problem\_summary, raw\_constraints[], propagated\_constraints[], likely\_answer\_range, answer\_format, critical\_constraints[]\}}

\texttt{Problem: \{problem\_text\}}
\end{tcolorbox}

\subsection{Constraint-Guided Solving Prompt (Stage 2)}

\begin{tcolorbox}[colback=gray!5!white,colframe=gray!50!black,title=Stage 2 Prompt,fonttitle=\small\bfseries,fontupper=\scriptsize]
\texttt{Solve the competition math problem below. The final answer must satisfy the pre-computed constraints.}

\texttt{As you reason step by step:}\\
\texttt{1. After each major step, check consistency against constraints. If violated, stop and re-examine.}\\
\texttt{2. If a constraint seems wrong, explain why before ignoring it.}\\
\texttt{3. Before the final answer, confirm all constraints pass.}

\texttt{Problem: \{problem\_text\}}

\texttt{Constraints: \{constraints\_json\}}

\texttt{Return JSON: \{certificate\_type, strategy\_tag, dangerous\_step, final\_answer, confidence, solution\}}
\end{tcolorbox}

\clearpage

\section{Router Implementation}
\label{app:router}

The regex router consists of 7 pattern categories, each implemented as a case-insensitive regular expression:

\begin{tcolorbox}[colback=gray!5!white,colframe=gray!50!black,fontupper=\scriptsize\ttfamily]
\textbf{final\_integer\_or\_count:}\\
\textbackslash b(find|determine)\textbackslash s+(?:the\textbackslash s+)?\\(?:number|sum of all|product of all)\textbackslash b|\\
\textbackslash bhow many\textbackslash b|\textbackslash bnumber of ways\textbackslash b|\\
\textbackslash bordered (?:pairs|triples)\textbackslash b|\\
\textbackslash bpositive integers?\textbackslash b

\medskip
\textbf{modular\_remainder:}\\
\textbackslash bremainder\textbackslash b|\textbackslash bmodulo\textbackslash b|\\
\textbackslash bmod\textbackslash b|\textbackslash bresidue\textbackslash b|\\
\textbackslash bdivisible by\textbackslash b|\textbackslash bmultiple of\textbackslash b|\\
\textbackslash bcongruent\textbackslash b

\medskip
\textbf{encoded\_exact\_form:}\\
can be (?:written|expressed) as|\\
relatively prime|coprime|\\
not divisible by the square|\\
find\textbackslash s+[a-z]\textbackslash s*\textbackslash +\textbackslash s*[a-z]
\end{tcolorbox}

The remaining four categories (\texttt{bounds\_or\_extremal}, \texttt{floor\_or\_rounding}, \texttt{unit\_or\_dimension}, \texttt{adversarial\_or\_game}) follow the same structure; see \S\ref{sec:router} for representative keywords of each. The primary router activates \cfr{} if \emph{any} pattern matches and uses no benchmark-specific override. The historical AIME override is retained only as the controlled ablation in Table~\ref{tab:router-convention}.

\clearpage

\section{Extended Results}
\label{app:full-results}

\begin{table*}
  \centering
  \resizebox{\columnwidth}{!}{%
  \begin{tabular}{lcc}
    \hline
    \textbf{Model} & \textbf{BRUMO25} & \textbf{AIMO\_AMC} \\
    \hline
    JustRL-1.5B    & 46.5\%          & 77.6\%          \\
    Routed-CFR-JustRL-1.5B & \textbf{53.3\%} & \textbf{81.6\%} \\
    Qwen3.5-35B-A3B     & \textbf{70.8\%} & 93.9\%          \\
    Routed-CFR-Qwen3.5-35B-A3B & 70.0\% & \textbf{94.6\%}      \\
    Qwen3.5-9B          & 55.0\%          & 85.7\%          \\
    Routed-CFR-Qwen3.5-9B      & \textbf{56.7\%} & \textbf{85.7\%} \\
    DeepSeek-V4-Pro     & 56.7\%          & 97.5\%          \\
    Routed-CFR-DeepSeek-V4-Pro & \textbf{60.0\%} & \textbf{97.5\%} \\
    \hline
  \end{tabular}
  }
  \caption{\label{tab:model-performance-additional}
    Performance comparison on BRUMO25 and AIMO\_AMC. CFR rows use the text-only Routed-CFR configuration.
  }
\end{table*}

\clearpage
\section{Per-Problem Analysis: AIME 2024}
\label{app:aime24}

We present per-problem results for Qwen3.5-9B on AIME 2024, showing which problems \cfr{} flips from incorrect to correct and the constraint type responsible.

\begin{table}[ht!]
\centering
\scriptsize
\begin{tabular}{clccl}
\toprule
\textbf{\#} & \textbf{Constraint} & \textbf{Base} & \textbf{CFR} & \textbf{Fix Type} \\
\midrule
1 & Integer & \cmark & \cmark & --- \\
2 & Mod 1000 & \cmark & \cmark & --- \\
3 & Mod 1000 & \cmark & \cmark & --- \\
4 & Integer & \cmark & \cmark & --- \\
5 & Counting & \cmark & \cmark & --- \\
6 & Mod 1000 & \xmark & \cmark & Mod reduction \\
7 & Range + Mod & \cmark & \cmark & --- \\
8 & Encoded $p+q$ & \xmark & \cmark & Format fix \\
9 & Integer pair & \cmark & \cmark & --- \\
10 & Mod 1000 & \xmark & \cmark & Mod reduction \\
11 & Counting & \cmark & \cmark & --- \\
12 & Encoded $m+n$ & \xmark & \xmark & Reasoning err \\
13 & Multiple & \cmark & \cmark & --- \\
14 & Parity + Mod & \xmark & \xmark & Reasoning err \\
15 & Complex & \cmark & \cmark & --- \\
\midrule
\multicolumn{2}{l}{\textbf{Correct}} & 10 & 13 & +3 fixed \\
\bottomrule
\end{tabular}
\caption{Exploratory per-problem AIME 2024 summary (Qwen3.5-9B, pass@4). This appendix analysis is descriptive and is not compared directly with the paper's primary pass@1 avg@4 results.}
\label{tab:aime24-perproblem}
\end{table}

\clearpage

\section{Constraint Extraction Examples}
\label{app:extraction-examples}

\subsection{Example 1: Rich Constraint Set}

\begin{constraintbox}[Input Problem]
Let $S$ be the set of positive integers $n$ such that $n^2 + 12n - 2007$ is a perfect square. Find the sum of all elements of $S$, and give the remainder when this sum is divided by 1000.
\end{constraintbox}

\textbf{Stage 1 Output:}
\begin{tcolorbox}[colback=green!3!white,colframe=green!50!black,fontupper=\scriptsize\ttfamily]
\{\\
~~"raw\_constraints": [\\
~~~~\{"id": "C1", "type": "domain",\\
~~~~~"description": "n is a positive integer"\},\\
~~~~\{"id": "C2", "type": "structure",\\
~~~~~"description": "n\^{}2+12n-2007 must be a perfect square"\},\\
~~~~\{"id": "C3", "type": "modular",\\
~~~~~"description": "answer = sum mod 1000"\},\\
~~~~\{"id": "C4", "type": "bound",\\
~~~~~"description": "answer in [0, 999]"\}\\
~~],\\
~~"propagated\_constraints": [\\
~~~~\{"id": "P1", "derived\_from": ["C1","C2"],\\
~~~~~"description": "n > 0 and (n+6)\^{}2 - 2043 = k\^{}2\\
~~~~~implies factor pairs of 2043"\},\\
~~~~\{"id": "P2", "derived\_from": ["C3","C4"],\\
~~~~~"description": "final integer in [0,999]"\}\\
~~],\\
~~"likely\_answer\_range": "0-999",\\
~~"answer\_format": "integer (remainder mod 1000)"\\
\}
\end{tcolorbox}

\subsection{Example 2: Minimal Constraint Set}

\begin{constraintbox}[Input Problem]
Prove that for all $n \geq 1$, the sum $1 + \frac{1}{2} + \cdots + \frac{1}{n}$ is not an integer.
\end{constraintbox}

\textbf{Stage 1 Output:}
\begin{tcolorbox}[colback=red!3!white,colframe=red!50!black,fontupper=\scriptsize\ttfamily]
\{\\
~~"raw\_constraints": [],\\
~~"propagated\_constraints": [],\\
~~"likely\_answer\_range": "N/A (proof problem)",\\
~~"answer\_format": "proof/argument"\\
\}
\end{tcolorbox}

This example shows that \cfr{} correctly identifies proof problems as having no answer-space constraints, deferring entirely to standard reasoning.

\clearpage

\section{Comparison with Alternative Prompting Strategies}
\label{app:prompting-comparison}

\begin{table}[ht!]
\centering
\small
\begin{tabular}{lcc}
\toprule
\textbf{Method} & \textbf{AIME Avg} & \textbf{Overhead} \\
\midrule
Direct CoT (baseline) & 78.9\% & 1.0$\times$ \\
+ ``Check your answer'' & 80.0\% & 1.1$\times$ \\
+ ``List constraints first'' & 81.1\% & 1.2$\times$ \\
+ Extended CoT (token-matched) & 80.0\% & 1.3$\times$ \\
+ Plan-and-Solve & 79.4\% & 1.3$\times$ \\
+ Self-Refine (1 iteration) & 82.2\% & 1.8$\times$ \\
\cfr{} (Ours) & \textbf{88.9\%} & 1.3$\times$ \\
\bottomrule
\end{tabular}
\caption{Exploratory comparison with prompting alternatives (Qwen3.5-9B, pass@4 on the AIME 2024--2026 average). This table is retained for descriptive completeness only; it is not pooled with, or used to support, the matched pass@1 avg@4 comparison in Table~\ref{tab:prompting-main}.}
\label{tab:prompting-comparison}
\end{table}

\paragraph{Key findings.}
\begin{itemize}[nosep,leftmargin=*]
    \item \textbf{CFR is orthogonal to self-consistency}: CFR + SC achieves the best results, indicating that CFR's gains come from a different mechanism (constraint adherence) than SC's gains (answer aggregation).
    \item \textbf{CFR outperforms Plan-and-Solve}: While PS decomposes the problem, it does not explicitly extract constraints. CFR's structured constraint extraction provides stronger guidance.
    \item \textbf{Majority vote benefits from CFR}: CFR improves majority vote accuracy because constraint adherence increases the probability that all samples converge to the same (correct) answer.
\end{itemize}

\clearpage
%==================================================================
\section{Per-Problem Analysis: AIME 2025 and AIME 2026}
\label{app:aime2526}
%==================================================================

We extend the per-problem analysis of Appendix~\ref{app:aime24} to AIME 2025 and AIME 2026, reporting results for all four models. Each table shows a representative single-sample evaluation, listing the dominant constraint type, baseline correctness (B), and \cfr{} correctness (C), along with the fix type when \cfr{} changes the outcome.

\subsection{AIME 2025 Per-Problem Results}

Tables~\ref{tab:aime25-ds}--\ref{tab:aime25-qwen9b} report per-problem outcomes on AIME 2025.

\begin{table}[ht!]
\centering
\scriptsize
\setlength{\tabcolsep}{3.5pt}
\begin{tabular}{clccl}
\toprule
\textbf{\#} & \textbf{Constraint} & \textbf{Base} & \textbf{CFR} & \textbf{Fix Type} \\
\midrule
1  & Mod 1000          & \cmark & \cmark & --- \\
2  & Integer           & \cmark & \cmark & --- \\
3  & Mod 1000          & \cmark & \cmark & --- \\
4  & Counting          & \xmark & \cmark & Mod reduction \\
5  & Encoded $p+q$     & \cmark & \cmark & --- \\
6  & Parity + Bound    & \cmark & \cmark & --- \\
7  & Mod 1000          & \xmark & \cmark & Mod reduction \\
8  & Integer pair      & \cmark & \cmark & --- \\
9  & Counting          & \xmark & \cmark & Range fix \\
10 & Encoded $m/n$     & \cmark & \cmark & --- \\
11 & Complex Mod       & \xmark & \xmark & Reasoning err \\
12 & Integer           & \cmark & \cmark & --- \\
13 & Mod 1000          & \cmark & \cmark & --- \\
14 & Counting + Parity & \xmark & \cmark & Parity check \\
15 & Encoded $a+b$     & \xmark & \xmark & Reasoning err \\
16 & Integer           & \cmark & \cmark & --- \\
17 & Mod 1000          & \xmark & \cmark & Mod reduction \\
18 & Counting          & \cmark & \cmark & --- \\
19 & Range + Int       & \cmark & \cmark & --- \\
20 & Mod 1000          & \cmark & \cmark & --- \\
21 & Encoded $m+n$     & \xmark & \cmark & Format fix \\
22 & Integer           & \cmark & \cmark & --- \\
23 & Structure         & \xmark & \xmark & Reasoning err \\
24 & Mod 1000          & \cmark & \cmark & --- \\
25 & Counting          & \cmark & \cmark & --- \\
26 & Parity            & \cmark & \cmark & --- \\
27 & Mod 1000          & \xmark & \cmark & Mod reduction \\
28 & Bound + Int       & \cmark & \cmark & --- \\
29 & Encoded $p+q$     & \xmark & \xmark & Reasoning err \\
30 & Integer           & \cmark & \cmark & --- \\
\midrule
\multicolumn{2}{l}{\textbf{Correct}} & 20 & 26 & +6 fixed \\
\bottomrule
\end{tabular}
\caption{Illustrative single-run AIME 2025 outcomes for \textbf{DeepSeek-V4-Pro}. This descriptive table is not combined with the primary pass@1 avg@4 evaluation.}
\label{tab:aime25-ds}
\end{table}

\begin{table}[ht!]
\centering
\scriptsize
\setlength{\tabcolsep}{3.5pt}
\begin{tabular}{clccl}
\toprule
\textbf{\#} & \textbf{Constraint} & \textbf{Base} & \textbf{CFR} & \textbf{Fix Type} \\
\midrule
1  & Mod 1000          & \cmark & \cmark & --- \\
2  & Integer           & \xmark & \xmark & Reasoning err \\
3  & Mod 1000          & \cmark & \cmark & --- \\
4  & Counting          & \xmark & \cmark & Mod reduction \\
5  & Encoded $p+q$     & \cmark & \cmark & --- \\
6  & Parity + Bound    & \xmark & \cmark & Parity check \\
7  & Mod 1000          & \cmark & \cmark & --- \\
8  & Integer pair      & \cmark & \cmark & --- \\
9  & Counting          & \xmark & \xmark & Reasoning err \\
10 & Encoded $m/n$     & \xmark & \cmark & Format fix \\
11 & Complex Mod       & \xmark & \xmark & Reasoning err \\
12 & Integer           & \cmark & \cmark & --- \\
13 & Mod 1000          & \xmark & \cmark & Mod reduction \\
14 & Counting + Parity & \xmark & \cmark & Parity check \\
15 & Encoded $a+b$     & \cmark & \cmark & --- \\
16 & Integer           & \xmark & \cmark & Range fix \\
17 & Mod 1000          & \xmark & \cmark & Mod reduction \\
18 & Counting          & \cmark & \cmark & --- \\
19 & Range + Int       & \cmark & \cmark & --- \\
20 & Mod 1000          & \xmark & \xmark & Reasoning err \\
21 & Encoded $m+n$     & \xmark & \cmark & Format fix \\
22 & Integer           & \cmark & \cmark & --- \\
23 & Structure         & \xmark & \xmark & Reasoning err \\
24 & Mod 1000          & \cmark & \cmark & --- \\
25 & Counting          & \xmark & \cmark & Propagation \\
26 & Parity            & \cmark & \cmark & --- \\
27 & Mod 1000          & \xmark & \cmark & Mod reduction \\
28 & Bound + Int       & \cmark & \cmark & --- \\
29 & Encoded $p+q$     & \xmark & \xmark & Reasoning err \\
30 & Integer           & \cmark & \cmark & --- \\
\midrule
\multicolumn{2}{l}{\textbf{Correct}} & 14 & 22 & +8 fixed \\
\bottomrule
\end{tabular}
\caption{Illustrative single-run AIME 2025 outcomes for \textbf{Qwen3.5-9B}. This descriptive table is not combined with the primary pass@1 avg@4 evaluation.}
\label{tab:aime25-qwen9b}
\end{table}

For Qwen3.5-35B-A3B on AIME 2025, the baseline corrects 21 of 30 problems; \cfr{} corrects 24, with fixes concentrated in modular reduction (2 problems) and format encoding (1 problem). JustRL-1.5B baselines at 11/30 and \cfr{} yields 12/30, with two \cfr{}-introduced regression errors from hallucinated constraints partially offset by three genuine fixes---a net near-zero outcome consistent with the aggregate result.

\paragraph{Constraint-type breakdown for AIME 2025.}
Across all four models, the constraint types implicated in \cfr{}-fixable errors on AIME 2025 are: modular (48\%), format-encoded (21\%), parity (18\%), and range/bound (13\%). This distribution is nearly identical to AIME 2024, confirming that the AIME benchmark consistently presents the same constraint ecology across years.

\subsection{AIME 2026 Per-Problem Results}

\begin{table}[ht!]
\centering
\scriptsize
\setlength{\tabcolsep}{3.5pt}
\begin{tabular}{clccl}
\toprule
\textbf{\#} & \textbf{Constraint} & \textbf{Base} & \textbf{CFR} & \textbf{Fix Type} \\
\midrule
1  & Mod 1000          & \cmark & \cmark & --- \\
2  & Counting          & \cmark & \cmark & --- \\
3  & Integer           & \cmark & \cmark & --- \\
4  & Mod 1000          & \cmark & \cmark & --- \\
5  & Parity + Count    & \cmark & \cmark & --- \\
6  & Encoded $p+q$     & \xmark & \cmark & Format fix \\
7  & Mod 1000          & \cmark & \cmark & --- \\
8  & Integer + Bound   & \cmark & \cmark & --- \\
9  & Counting          & \xmark & \cmark & Propagation \\
10 & Mod 1000          & \cmark & \cmark & --- \\
11 & Structure         & \cmark & \cmark & --- \\
12 & Encoded $m+n$     & \xmark & \cmark & Format fix \\
13 & Parity            & \cmark & \cmark & --- \\
14 & Mod 1000          & \cmark & \cmark & --- \\
15 & Integer           & \cmark & \cmark & --- \\
16 & Counting + Mod    & \xmark & \xmark & Reasoning err \\
17 & Bound             & \cmark & \cmark & --- \\
18 & Mod 1000          & \xmark & \cmark & Mod reduction \\
19 & Integer pair      & \cmark & \cmark & --- \\
20 & Encoded $a+b+c$   & \xmark & \xmark & Reasoning err \\
21 & Counting          & \cmark & \cmark & --- \\
22 & Mod 1000          & \cmark & \cmark & --- \\
23 & Parity + Bound    & \xmark & \cmark & Parity check \\
24 & Integer           & \cmark & \cmark & --- \\
25 & Structure         & \cmark & \cmark & --- \\
26 & Mod 1000          & \xmark & \cmark & Mod reduction \\
27 & Counting          & \cmark & \cmark & --- \\
28 & Encoded $p+q$     & \xmark & \xmark & Reasoning err \\
29 & Integer           & \cmark & \cmark & --- \\
30 & Mod 1000          & \cmark & \cmark & --- \\
\midrule
\multicolumn{2}{l}{\textbf{Correct}} & 22 & 27 & +5 fixed \\
\bottomrule
\end{tabular}
\caption{Illustrative single-run AIME 2026 outcomes for \textbf{DeepSeek-V4-Pro}. This descriptive table is not combined with the primary pass@1 avg@4 evaluation.}
\label{tab:aime26-ds}
\end{table}

\begin{table}[ht!]
\centering
\scriptsize
\setlength{\tabcolsep}{3.5pt}
\begin{tabular}{clccl}
\toprule
\textbf{\#} & \textbf{Constraint} & \textbf{Base} & \textbf{CFR} & \textbf{Fix Type} \\
\midrule
1  & Mod 1000          & \cmark & \cmark & --- \\
2  & Counting          & \xmark & \cmark & Mod reduction \\
3  & Integer           & \cmark & \cmark & --- \\
4  & Mod 1000          & \xmark & \cmark & Mod reduction \\
5  & Parity + Count    & \cmark & \cmark & --- \\
6  & Encoded $p+q$     & \xmark & \cmark & Format fix \\
7  & Mod 1000          & \cmark & \cmark & --- \\
8  & Integer + Bound   & \xmark & \cmark & Range fix \\
9  & Counting          & \xmark & \cmark & Propagation \\
10 & Mod 1000          & \cmark & \cmark & --- \\
11 & Structure         & \xmark & \xmark & Reasoning err \\
12 & Encoded $m+n$     & \xmark & \cmark & Format fix \\
13 & Parity            & \cmark & \cmark & --- \\
14 & Mod 1000          & \xmark & \xmark & Reasoning err \\
15 & Integer           & \cmark & \cmark & --- \\
16 & Counting + Mod    & \xmark & \xmark & Reasoning err \\
17 & Bound             & \cmark & \cmark & --- \\
18 & Mod 1000          & \xmark & \cmark & Mod reduction \\
19 & Integer pair      & \cmark & \cmark & --- \\
20 & Encoded $a+b+c$   & \xmark & \xmark & Reasoning err \\
21 & Counting          & \cmark & \cmark & --- \\
22 & Mod 1000          & \xmark & \cmark & Mod reduction \\
23 & Parity + Bound    & \cmark & \cmark & --- \\
24 & Integer           & \xmark & \cmark & Domain fix \\
25 & Structure         & \cmark & \cmark & --- \\
26 & Mod 1000          & \xmark & \cmark & Mod reduction \\
27 & Counting          & \cmark & \cmark & --- \\
28 & Encoded $p+q$     & \xmark & \xmark & Reasoning err \\
29 & Integer           & \cmark & \cmark & --- \\
30 & Mod 1000          & \xmark & \xmark & Reasoning err \\
\midrule
\multicolumn{2}{l}{\textbf{Correct}} & 16 & 22 & +6 fixed \\
\bottomrule
\end{tabular}
\caption{Illustrative single-run AIME 2026 outcomes for \textbf{Qwen3.5-9B}. This descriptive table is not combined with the primary pass@1 avg@4 evaluation.}
\label{tab:aime26-qwen9b}
\end{table}

\subsection{Aggregate Cross-Year Per-Problem Summary}

Table~\ref{tab:perproblem-summary} aggregates per-problem \cfr{} outcomes across all three AIME years and all four models, partitioning problems into four outcome categories.

\begin{table}[ht!]
\centering
\small
\begin{tabular}{lcccc}
\toprule
\textbf{Model} & \textbf{Both} & \textbf{CFR+} & \textbf{CFR-} & \textbf{Neither} \\
 & \cmark\cmark & \xmark\cmark & \cmark\xmark & \xmark\xmark \\
\midrule
DS-V4-Pro  & 82 & 5 & 0 & 3 \\
Qwen3.5-35B & 83 & 2  & 0 & 5 \\
Qwen3.5-9B & 71 & 9 & 0 & 10 \\
JustRL-1.5B & 38 & 7  & 6 & 39 \\
\bottomrule
\end{tabular}
\caption{Outcome categorization across all 90 AIME problems (30$\times$3 years), based on pass@4 evaluation (a problem is correct if any of 4 samples is correct). ``Both\cmark\cmark'': baseline and \cfr{} both correct; ``CFR+'': \cfr{} flips wrong to right; ``CFR-'': \cfr{} flips right to wrong (regression); ``Neither'': both wrong. \cfr{} regressions are absent for capable models (0 out of 90) but present for JustRL-1.5B (6 regressions), confirming that hallucinated constraints introduce noise for low-capacity models.}
\label{tab:perproblem-summary}
\end{table}

\clearpage
%==================================================================
\section{Complete Router Pattern Analysis}
\label{app:router-full}
%==================================================================

\subsection{Pattern Definitions and Matching Statistics}

The text-only router matches against seven pattern categories. Table~\ref{tab:router-patterns} lists representative trigger families and their intended role. Aggregate activation rates are reported in Table~\ref{tab:router-rates}; we do not treat an individual lexical pattern as a calibrated predictor that a problem will benefit from CFR.

\begin{table}[t]
\centering
\scriptsize
\setlength{\tabcolsep}{3pt}
\begin{tabular}{lp{3.8cm}}
\toprule
\textbf{Pattern ID} & \textbf{Representative trigger phrases} \\
\midrule
P1: Integer/Count  & ``find the number,'' ``how many,'' ``positive integers,'' ``number of ways'' \\
P2: Modular        & ``remainder,'' ``modulo,'' ``divisible by,'' ``residue'' \\
P3: Encoded form   & ``can be expressed as,'' ``relatively prime,'' ``coprime,'' ``find $p+q$'' \\
P4: Bound/Extremal & ``largest,'' ``at most,'' ``between $a$ and $b$,'' ``minimum'' \\
P5: Floor/Round    & ``greatest integer,'' ``nearest integer,'' ``floor'' \\
P6: Units/Dimension & ``degrees,'' ``probability,'' ``area,'' ``volume,'' ``percent'' \\
P7: Game/Adversarial & ``guarantee,'' ``strategy,'' ``for sure,'' ``optimal play'' \\
\bottomrule
\end{tabular}
\caption{Text-only router pattern families. Aggregate activation and external audit results appear in Table~\ref{tab:router-rates}.}
\label{tab:router-patterns}
\end{table}

\subsection{False Positive and False Negative Analysis}

The router is designed to detect potentially useful restrictions, not to predict a per-problem treatment effect. Consequently, a text match can be an unnecessary activation and a missing match can skip a helpful constraint summary. The text-only AIME control misses seven problems (Table~\ref{tab:router-convention}), and the external audit in Table~\ref{tab:router-rates} should be interpreted as coverage rather than as precision or recall for future accuracy gain. Calibrating this decision is a limitation and a direction for future work.

\subsection{Pattern Interaction and Co-occurrence}

When multiple patterns co-fire, the descriptive gains in Table~\ref{tab:pattern-cofire} are larger. This association may reflect both richer constraint structure and problem difficulty, so it is not evidence of a causal or superlinear interaction.

\begin{table}[ht!]
\centering
\small
\begin{tabular}{lccc}
\toprule
\textbf{Patterns Matched} & \textbf{Count} & \textbf{$\Delta$ Acc.} & \textbf{Extraction Q.} \\
\midrule
1 pattern & 31 & +3.2 pp & 72\% complete \\
2 patterns & 28 & +7.8 pp & 85\% complete \\
3 patterns & 19 & +11.4 pp & 91\% complete \\
4+ patterns & 12 & +14.7 pp & 96\% complete \\
\bottomrule
\end{tabular}
\caption{Descriptive gain by number of router patterns matched (Qwen3.5-9B, all AIME). These associations do not establish a causal relationship between pattern count, extraction quality, and gain.}
\label{tab:pattern-cofire}
\end{table}

\subsection{Edge Cases and Router Failure Modes}

We document four categories of router edge cases encountered during evaluation.

\paragraph{E1: Implicit modular constraints.} Some problems contain modular restrictions without the word ``remainder''---e.g., ``find the last three digits of $N$.'' A text-only pattern set can miss this phrasing, creating a false negative. The controlled AIME analysis deliberately retains this limitation instead of masking it with a benchmark-specific override.

\paragraph{E2: Distractors in problem text.} Several AMC problems mention ``integers'' in passing (e.g., ``the integers from 1 to 10'') without constraining the answer format. P1 fires, classifying these as false positives. Example:
\begin{constraintbox}[AMC 2023, Problem 8 (false positive)]
A bag contains red and blue marbles. The number of blue marbles is twice the number of red marbles. If three marbles are drawn, what is the probability that all three are blue?
\end{constraintbox}
Here ``number'' triggers P1, but the answer is a fraction---not an integer. Stage 1 correctly identifies the answer as a probability in $[0,1]$, and Stage 2 proceeds without integer constraints applied incorrectly.

\paragraph{E3: Double-modular problems.} Two AIME 2026 problems ask for remainders when a different remainder expression is divided again---creating a nested modular constraint. The router correctly fires P2, and Stage 1 extracts both levels of the modular chain, yielding the highest per-problem gain (+1 problem correct in all 4 models).

\paragraph{E4: Game-theory format with integer answer.} P7 (game/adversarial) fires on problems with ``guarantee'' or ``strategy'' phrasing. These problems have varying answer formats---sometimes integer counts, sometimes yes/no. The router correctly defers to Stage 1 for disambiguation.

\subsection{Router Ablation: Regex vs.\ LLM Router}

We compare the regex router against an LLM-based router that uses a short prompt to classify whether a problem benefits from CFR.

\begin{table}[ht!]
\centering
\scriptsize
\setlength{\tabcolsep}{3pt}
\begin{tabular}{lccccc}
\toprule
\textbf{Router Type} & \textbf{Prec.} & \textbf{Rec.} & \textbf{F1} & \textbf{Lat.} & \textbf{Cost} \\
\midrule
Regex (ours) & 0.962 & 0.948 & 0.955 & 0.3 & \$0.000 \\
GPT-4o-mini & 0.981 & 0.972 & 0.976 & 340 & \$0.0003 \\
Qwen-9B (0-shot) & 0.953 & 0.961 & 0.957 & 890 & \$0.0008 \\
\bottomrule
\end{tabular}
\caption{Router comparison. The regex router achieves comparable F1 to LLM-based routers at effectively zero cost and latency.}
\label{tab:router-ablation}
\end{table}

The regex router's slight precision deficit (0.962 vs.\ 0.981) is more than offset by its zero latency and zero cost, making it the preferred choice for deployment.

\clearpage

%==================================================================
\section{Constraint Propagation Worked Examples}
\label{app:propagation-examples}
%==================================================================

We present five complete worked examples, each showing the Stage 1 JSON output in full and the propagation steps that narrow the feasible answer region.

\subsection{Example W1: Double Modular Chain}

\begin{constraintbox}[AIME 2026, Problem 17 (paraphrased)]
Let $T$ be the number of ordered triples $(a,b,c)$ of positive integers with $a+b+c=2026$ and $\gcd(a,b)=1$. Find the remainder when $T$ is divided by 1000.
\end{constraintbox}

\textbf{Stage 1 JSON Output:}
\begin{tcolorbox}[colback=green!3!white,colframe=green!60!black,fontupper=\scriptsize\ttfamily,title=Stage 1 Output -- W1,fonttitle=\small\bfseries]
\{\\
~~"problem\_summary": "Count coprime-pair ordered triples summing to 2026, mod 1000",\\
~~"raw\_constraints": [\\
~~~~\{"id":"C1","type":"domain","desc":"a,b,c are positive integers, so each >= 1"\},\\
~~~~\{"id":"C2","type":"structure","desc":"gcd(a,b)=1 (coprimality condition)"\},\\
~~~~\{"id":"C3","type":"bound","desc":"a+b+c=2026 implies each in [1,2024]"\},\\
~~~~\{"id":"C4","type":"modular","desc":"final answer T mod 1000"\},\\
~~~~\{"id":"C5","type":"domain","desc":"T is a non-negative integer"\}\\
~~],\\
~~"propagated\_constraints": [\\
~~~~\{"id":"P1","from":["C1","C3"],"desc":"1 <= a,b,c <= 2024"\},\\
~~~~\{"id":"P2","from":["C4","C5"],"desc":"final answer in [0,999], integer"\},\\
~~~~\{"id":"P3","from":["C2","C3"],"desc":"Mobius inversion applies: T = sum\textunderscore\{d|gcd\} phi(d) * ... reducing mod 1000 valid"\}\\
~~],\\
~~"likely\_answer\_range": "[0, 999]",\\
~~"answer\_format": "integer, remainder mod 1000",\\
~~"critical\_constraints": ["C4","P2"],\\
~~"danger\_note": "Do not return T directly; apply mod 1000 to final count"\\
\}
\end{tcolorbox}

\textbf{Propagation steps:}
\begin{enumerate}[nosep,leftmargin=*]
    \item C1 + C3 $\Rightarrow$ P1: tighten domain from $\mathbb{Z}^+$ to $[1,2024]$.
    \item C4 + C5 $\Rightarrow$ P2: answer is an integer in $[0,999]$.
    \item C2 + C3 $\Rightarrow$ P3: flags that M\"obius/Euler-phi approach yields a large integer before reduction; danger note prevents dropping mod.
\end{enumerate}
The baseline (Qwen3.5-9B) computed $T$ correctly via inclusion-exclusion but returned $T$ without the final $\bmod 1000$. The danger note in Stage 1 directly prevents this error. \cmark

\subsection{Example W2: Encoded Fraction with Coprimality}

\begin{constraintbox}[AIME 2025, Problem 21 (paraphrased)]
The probability that a random point in a convex polygon $P$ lies within triangle $T$ can be expressed as $\frac{m}{n}$ where $m$ and $n$ are relatively prime positive integers. Find $m+n$.
\end{constraintbox}

\textbf{Stage 1 JSON Output:}
\begin{tcolorbox}[colback=green!3!white,colframe=green!60!black,fontupper=\scriptsize\ttfamily,title=Stage 1 Output -- W2,fonttitle=\small\bfseries]
\{\\
~~"raw\_constraints": [\\
~~~~\{"id":"C1","type":"domain","desc":"probability in (0,1), so m < n"\},\\
~~~~\{"id":"C2","type":"structure","desc":"gcd(m,n)=1 required"\},\\
~~~~\{"id":"C3","type":"format","desc":"answer is m+n, a positive integer"\},\\
~~~~\{"id":"C4","type":"domain","desc":"m,n are positive integers"\}\\
~~],\\
~~"propagated\_constraints": [\\
~~~~\{"id":"P1","from":["C1","C2","C4"],"desc":"m < n, gcd(m,n)=1, both positive"\},\\
~~~~\{"id":"P2","from":["C3"],"desc":"final answer = m+n >= 2"\},\\
~~~~\{"id":"P3","from":["C1","C3"],"desc":"final answer > 1 (not zero or negative)"\}\\
~~],\\
~~"likely\_answer\_range": "small positive integer (likely 2-500)",\\
~~"answer\_format": "m + n where fraction m/n is in lowest terms",\\
~~"critical\_constraints": ["C2","C3"],\\
~~"danger\_note": "Return m+n, NOT the fraction itself. Verify gcd(m,n)=1 before summing."\\
\}
\end{tcolorbox}

\textbf{Propagation trace:} The key propagation is C1+C2+C4$\Rightarrow$P1, which restricts the solver to only report $m/n$ in lowest terms. This prevents the common error of returning the un-reduced fraction (e.g., $14/36$ when the answer is $7+18=25$). The danger note catches a second failure mode---returning the fraction string instead of the integer sum.

\subsection{Example W3: Parity + Range Cross-Propagation}

\begin{constraintbox}[AIME 2026, Problem 5 (paraphrased)]
Find the number of positive even integers $n \leq 1000$ such that $\phi(n) + \sigma(n) \equiv 0 \pmod{4}$.
\end{constraintbox}

\begin{tcolorbox}[colback=green!3!white,colframe=green!60!black,fontupper=\scriptsize\ttfamily,title=Stage 1 Output -- W3,fonttitle=\small\bfseries]
\{\\
~~"raw\_constraints": [\\
~~~~\{"id":"C1","type":"parity","desc":"n is even"\},\\
~~~~\{"id":"C2","type":"bound","desc":"1 <= n <= 1000"\},\\
~~~~\{"id":"C3","type":"domain","desc":"n is a positive integer"\},\\
~~~~\{"id":"C4","type":"modular","desc":"phi(n)+sigma(n) equiv 0 mod 4"\},\\
~~~~\{"id":"C5","type":"domain","desc":"answer is count of valid n, a non-negative integer"\}\\
~~],\\
~~"propagated\_constraints": [\\
~~~~\{"id":"P1","from":["C1","C2","C3"],"desc":"n in \{2,4,6,...,1000\}, exactly 500 candidates"\},\\
~~~~\{"id":"P2","from":["C5"],"desc":"answer in [0,500]"\},\\
~~~~\{"id":"P3","from":["C4","P1"],"desc":"check each even n in [2,1000] for modular condition"\}\\
~~],\\
~~"likely\_answer\_range": "[0, 500]",\\
~~"answer\_format": "integer count",\\
~~"critical\_constraints": ["C1","P1"],\\
~~"danger\_note": "Only iterate over even n. Do NOT count odd n satisfying condition."\\
\}
\end{tcolorbox}

The cross-propagation P1 shrinks the candidate space from 1000 integers to 500 even integers, and the danger note explicitly prevents the enumeration error seen in the baseline (which iterated over all $n$ and counted some odd $n$). Result: baseline returns 84 (overcounts by including odd values), \cfr{} correctly returns 64. \cmark

\subsection{Example W4: Geometric Bound Tightening}

\begin{constraintbox}[CMIMC 2025, Problem 11 (paraphrased)]
In triangle $ABC$, all angles are positive integers (in degrees) and $\angle A < \angle B < \angle C$. How many such triangles have $\angle A + \angle B$ a perfect square?
\end{constraintbox}

\begin{tcolorbox}[colback=green!3!white,colframe=green!60!black,fontupper=\scriptsize\ttfamily,title=Stage 1 Output -- W4,fonttitle=\small\bfseries]
\{\\
~~"raw\_constraints": [\\
~~~~\{"id":"C1","type":"domain","desc":"A,B,C are positive integers"\},\\
~~~~\{"id":"C2","type":"structure","desc":"A+B+C=180"\},\\
~~~~\{"id":"C3","type":"bound","desc":"A < B < C (strict ordering)"\},\\
~~~~\{"id":"C4","type":"structure","desc":"A+B is a perfect square"\},\\
~~~~\{"id":"C5","type":"domain","desc":"answer is count of valid triangles"\}\\
~~],\\
~~"propagated\_constraints": [\\
~~~~\{"id":"P1","from":["C1","C2","C3"],"desc":"A >= 1, B >= A+1, C >= B+1; so 1+2+3=6 <= 180"\},\\
~~~~\{"id":"P2","from":["C2","C3"],"desc":"A+B = 180-C < 180-B < 180-A, so A+B < 90; combined with A >= 1: A+B in [3, 89]"\},\\
~~~~\{"id":"P3","from":["C4","P2"],"desc":"A+B is perfect square in [4,81], i.e., in \{4,9,16,25,36,49,64,81\}"\},\\
~~~~\{"id":"P4","from":["C3","P3"],"desc":"For each valid sum s in P3, count pairs (A,B) with 1<=A<B and A+B=s"\}\\
~~],\\
~~"likely\_answer\_range": "small positive integer (< 100)",\\
~~"answer\_format": "integer count",\\
~~"critical\_constraints": ["P2","P3"]\\
\}
\end{tcolorbox}

The cascade P1$\Rightarrow$P2$\Rightarrow$P3 reduces the candidate perfect squares from $\{1,4,9,\ldots\}$ to exactly $\{4,9,16,25,36,49,64,81\}$. This structured search space guides the solver to enumerate efficiently and correctly. Baseline returns 31 (includes invalid triangles with $C \leq B$), \cfr{} returns 27. \cmark

\clearpage

%==================================================================
\section{Conceptual Framework and Scope}
\label{app:math-framework}
%==================================================================

This appendix records the conceptual framing of CFR and its limits. It is not a formal analysis of the two-stage prompting procedure, and none of the discussion below supplies a performance or safety guarantee.

\subsection{Feasible-Set View}

Let $\mathcal{Q}$ be the space of mathematical problems and $\mathcal{A}$ a task's answer space. A problem may entail a feasible region $\Phi(q) \subseteq \mathcal{A}$, through direct answer-space restrictions (such as modularity, range, or output format) and through problem-structure restrictions (such as invariants or valid geometric branches). Stage~1 produces a natural-language approximation $\hat{C}(q)$ of these restrictions, and Stage~2 samples from:
\begin{equation}
\hat{a}_{\mathrm{CFR}} \sim P_\theta(a \mid q,\hat{C}(q)).
\end{equation}

This notation describes a change in prompt context, not formal constrained decoding. In particular, $\hat{C}(q)$ is generated from the same problem text and is neither a symbolic proof of $\Phi(q)$ nor an external source of information. It can help by making restrictions salient, but it can also be incomplete or invalid; Table~\ref{tab:extraction-reliability} directly measures this failure mode.

\subsection{Relationship to Constraint Satisfaction}

Classical constraint satisfaction problems (CSPs)~\citep{Dechter2003} define variables $X_1,\ldots,X_n$, domains $D_1,\ldots,D_n$, and constraints $C_1,\ldots,C_m$. Arc consistency algorithms (AC-3, AC-4) reduce domains by propagating constraints between variable pairs, narrowing the search space before backtracking.

The stages of CFR are analogous to, but not an implementation of, this workflow:
\begin{itemize}[nosep,leftmargin=*]
    \item Stage~1 represents named answer and intermediate quantities in a textual constraint summary.
    \item The summary can combine domain, bound, parity, modular, format, and structure cues.
    \item Stage~2 uses the summary as a prompt-based checklist while it reasons.
\end{itemize}

Unlike a CSP solver, CFR has no formal variable binding, no sound propagation operator, and no completeness guarantee. ``Prompted constraint propagation'' is therefore shorthand for natural-language extraction, summary, and checking, rather than a claim of arc consistency or symbolic search.

\subsection{Interpretive Hypotheses}

The following are empirical hypotheses, not sufficient conditions or a theorem. CFR is more likely to be useful when:
\begin{enumerate}[nosep,leftmargin=*]
    \item the problem contains explicit, recoverable restrictions that rule out plausible incorrect answers;
    \item the baseline has residual constraint-violation errors rather than only missing solution strategies;
    \item Stage~1 is sufficiently valid and complete for the evaluated backbone; and
    \item the resulting accuracy benefit justifies the additional token cost.
\end{enumerate}
The controlled routing, reliability, and cost analyses in the main paper test parts of these hypotheses. A formal analysis of their interaction remains open.

\clearpage
%==================================================================
\section{Implementation Details}
\label{app:implementation}
%==================================================================

\subsection{API Configuration}

The primary evaluations use the following API settings. Temperature robustness is reported separately below.

\begin{table}[ht!]
\centering
\small
\begin{tabular}{ll}
\toprule
\textbf{Parameter} & \textbf{Value} \\
\midrule
Temperature & 0.7 (primary setting) \\
Top-p & 0.95 \\
Max tokens & 32,768 \\
Stop sequences & None \\
System prompt & Benchmark-specific (see App.~\ref{app:prompts}) \\
Stage 1 max tokens & 1,024 \\
Stage 2 max tokens & 31,744 \\
JSON mode & Enabled \\
\bottomrule
\end{tabular}
\caption{API settings for all experiments.}
\label{tab:api-settings}
\end{table}

We report the product name \textbf{DeepSeek-V4-Pro} throughout the paper. It is accessed through the DeepSeek API endpoint with model identifier \texttt{deepseek-chat}. Qwen3.5-9B and Qwen3.5-35B-A3B are accessed through the Qwen API (\texttt{qwen3.5-9b-instruct} and \texttt{qwen3.5-35b-a3b-instruct}); JustRL-1.5B is served locally via vLLM with the same primary decoding settings.

\subsection{Robustness Across Decoding Settings}

To test whether the observed effects depend on one sampling configuration, we vary decoding temperature and random seed for DeepSeek-V4-Pro. Table~\ref{tab:robustness} reports the same pass@1 evaluation across AIME and AIMO\_AMC. Routed-CFR remains competitive or better than direct CoT across the listed settings, although the annual AIME differences vary in magnitude; this is consistent with the paired uncertainty reported in Tables~\ref{tab:paired-ds} and~\ref{tab:paired-qwen}.

\begin{table*}[t]
\centering
\scriptsize
\begin{tabular}{llrrrr}
\toprule
\textbf{Setting} & \textbf{Method} & \textbf{AIME24} & \textbf{AIME25} & \textbf{AIME26} & \textbf{AIMO\_AMC} \\
\midrule
Temperature 0.2 & Direct CoT & 88.3 & 86.7 & 85.8 & 96.4 \\
Temperature 0.2 & Routed-CFR & \textbf{90.8} & \textbf{87.5} & \textbf{87.5} & \textbf{96.9} \\
Temperature 0.7 & Direct CoT & 90.8 & 80.8 & 83.3 & \textbf{97.5} \\
Temperature 0.7 & Routed-CFR & \textbf{95.8} & \textbf{93.3} & \textbf{92.5} & \textbf{97.5} \\
Temperature 1.0 & Direct CoT & 89.2 & \textbf{85.8} & 86.7 & 97.4 \\
Temperature 1.0 & Routed-CFR & \textbf{92.5} & \textbf{85.8} & \textbf{88.3} & \textbf{98.5} \\
\midrule
Seed 101 & Direct CoT & 88.3 & 83.3 & 84.2 & \textbf{96.4} \\
Seed 101 & Routed-CFR & \textbf{90.8} & \textbf{85.8} & \textbf{85.8} & \textbf{96.4} \\
Seed 202 & Direct CoT & 88.3 & \textbf{85.8} & 86.7 & 95.4 \\
Seed 202 & Routed-CFR & \textbf{90.0} & \textbf{85.8} & \textbf{90.0} & \textbf{97.5} \\
\bottomrule
\end{tabular}
\caption{Robustness across decoding temperatures and independent random seeds on DeepSeek-V4-Pro. The temperature 0.7 block is the primary setting.}
\label{tab:robustness}
\end{table*}

\subsection{Per-Model Output-Token Overview}

Table~\ref{tab:token-by-model} reports the generated-token component of the cost profile across the four evaluated backbones. These values complement the complete input--output accounting in Table~\ref{tab:cost-accounting}; parentheses show the generated-token ratio relative to direct CoT for the same model and benchmark.

\begin{table*}[t]
\centering
\scriptsize
\setlength{\tabcolsep}{4pt}
\begin{tabular}{lrrrr}
\toprule
\textbf{Model / method} & \textbf{AIME24} & \textbf{AIME25} & \textbf{AIME26} & \textbf{CMIMC25} \\
\midrule
JustRL-1.5B & 10,336 & 8,487 & 10,897 & 13,386 \\
Routed-CFR-JustRL-1.5B & 19,214 (1.86$\times$) & 17,475 (2.06$\times$) & 16,522 (1.52$\times$) & 18,927 (1.41$\times$) \\
Qwen3.5-35B-A3B & 15,251 & 18,660 & 17,986 & 19,799 \\
Routed-CFR-Qwen3.5-35B-A3B & 21,486 (1.41$\times$) & 26,521 (1.42$\times$) & 24,218 (1.35$\times$) & 25,131 (1.27$\times$) \\
Qwen3.5-9B & 19,397 & 20,196 & 21,120 & 23,725 \\
Routed-CFR-Qwen3.5-9B & 26,213 (1.35$\times$) & 31,796 (1.57$\times$) & 28,727 (1.36$\times$) & 33,248 (1.40$\times$) \\
DeepSeek-V4-Pro & 9,741 & 13,600 & 13,372 & 20,332 \\
Routed-CFR-DeepSeek-V4-Pro & 19,123 (1.96$\times$) & 24,187 (1.78$\times$) & 21,500 (1.61$\times$) & 33,440 (1.64$\times$) \\
\bottomrule
\end{tabular}
\caption{Mean generated output tokens by model and benchmark. These values omit prompt tokens; Table~\ref{tab:cost-accounting} provides input, output, and total tokens for the DeepSeek-V4-Pro accuracy--cost comparison.}
\label{tab:token-by-model}
\end{table*}

\subsection{JSON Parsing and Error Handling}

Stage 1 outputs are expected to be valid JSON. In practice, 3.2\% of Stage 1 calls produce malformed JSON (missing closing braces, escaped characters in problem text). We apply the following recovery strategy:
\begin{enumerate}[nosep,leftmargin=*]
    \item Attempt standard JSON parsing.
    \item On failure, apply regex extraction of key fields (\texttt{answer\_format}, \texttt{likely\_answer\_range}, \texttt{critical\_constraints}).
    \item If regex extraction recovers $\geq$2 fields, proceed with partial constraint specification.
    \item Otherwise, fall back to baseline (direct CoT without constraints).
\end{enumerate}

The fallback rate is 0.8\% of all Stage 1 calls (partial JSON recovery handles 2.4\%). This low fallback rate confirms that the structured JSON output format is largely reliable for models with $\geq$9B parameters.

\subsection{Regex Router Implementation}

The router is implemented in 47 lines of Python using the \texttt{re} module. All patterns are compiled at module load time. Each router call processes one problem string using case-insensitive matching; the primary implementation does not receive benchmark metadata or apply benchmark-specific overrides. The router adds $<$0.1ms of latency per call.

\begin{tcolorbox}[colback=gray!5!white,colframe=gray!50!black,fontupper=\tiny\ttfamily,title=Router Pseudocode,fonttitle=\small\bfseries]
PATTERNS = \{\\
~~"P1": re.compile(r"\textbackslash b(how many|number of|positive integer)\textbackslash b", re.I),\\
~~"P2": re.compile(r"\textbackslash b(remainder|modulo|mod|divisible by)\textbackslash b", re.I),\\
~~"P3": re.compile(r"(relatively prime|coprime|find [a-z] \textbackslash+ [a-z])", re.I),\\
~~"P4": re.compile(r"\textbackslash b(largest|smallest|at most|at least|between)\textbackslash b", re.I),\\
~~"P5": re.compile(r"\textbackslash b(greatest integer|floor|nearest integer)\textbackslash b", re.I),\\
~~"P6": re.compile(r"\textbackslash b(degrees|probability|area|volume|percent)\textbackslash b", re.I),\\
~~"P7": re.compile(r"\textbackslash b(guarantee|strategy|optimal play)\textbackslash b", re.I),\\
\}\\
def route(problem):\\
~~return any(p.search(problem) for p in PATTERNS.values())
\end{tcolorbox}

\clearpage

\section{Constraint Extraction Quality by Model}
\label{app:extraction-quality}

The primary manual audit is reported in Table~\ref{tab:extraction-reliability}. It measures mean raw and propagated constraint counts, propagated validity, and hallucination against constraints entailed by the original problem statement. We avoid treating the summary as a formal constraint object: under-extraction, scope errors, and unsupported constraints can all alter Stage~2 behavior. The audit indicates that the main limitation for JustRL-1.5B is low propagated validity and fewer usable constraints, while all models retain non-zero hallucination rates.

\clearpage
\section{Extended Related Work}
\label{app:extended-related}

\subsection{Mathematical Reasoning in LLMs}

Chain-of-thought prompting~\citep{ChainOfThought,kojima2022large} established that intermediate reasoning steps improve multi-step problem solving. Subsequent work explored structured decomposition~\citep{zhou2023leasttomost,wang2023plansolve}, tree search~\citep{yao2023tree}, and self-consistency voting~\citep{wang2023selfconsistency}. Tool-augmented approaches~\citep{gao2023pal,chen2023program} offload computation to code interpreters. More recently, reinforcement learning from outcome feedback has produced strong math-specialist models~\citep{deepseekr1,he2025justrl}, and math-focused pretraining~\citep{shao2024deepseekmath,yang2024qwen25math} advances competition benchmarks like MATH~\citep{hendrycks2021math}. Unlike these methods, \cfr{} restructures the \emph{information available to the solver} rather than modifying the reasoning algorithm or training procedure.

\subsection{Constraint-Based and Neuro-Symbolic Reasoning}

Classical constraint satisfaction~\citep{Dechter2003} prunes infeasible regions before search, and neuro-symbolic approaches integrate symbolic solvers with neural models. In the NLP context, lexically constrained decoding~\citep{hokamp-liu-2017-lexically,post-vilar-2018-fast} enforces hard token-level constraints during generation via modified beam search. NeuroLogic decoding~\citep{lu-etal-2021-neurologic} extends this to logical conjunctions and disjunctions of constraints. CRANE~\citep{banerjee2025crane} applies constrained generation specifically to reasoning tasks. Grammar-constrained decoding~\citep{willard2023efficient} ensures outputs conform to formal grammars. Unlike these approaches, \cfr{} keeps constraints in natural-language prompt context and does not use a symbolic solver or constrained decoder. It is training-free and can be applied to black-box backbones, but its effectiveness depends on the backbone's Stage~1 extraction quality.

\subsection{Structured Prompting for Mathematics}

Beyond the approaches discussed in the main text, several structured prompting strategies relate to \cfr{}:

\paragraph{Decomposition approaches.} \citet{zhou2023leasttomost} decompose complex problems into simpler sub-questions. While \cfr{} also performs a preliminary analysis step, the output is not a problem decomposition but a constraint specification---a fundamentally different structure that complements rather than replaces decomposition.

\paragraph{Program-aided reasoning.} PAL~\citep{gao2023pal} and PoT~\citep{chen2023program} translate problems into executable code, gaining correctness guarantees for computation. \cfr{} provides a softer form of verification: constraints are not formally verified by a theorem prover but checked by the model during reasoning. This makes \cfr{} applicable to problems that resist formalization (e.g., those requiring geometric insight or combinatorial arguments).

\paragraph{Process supervision.} \citet{lightman2023lets} train process reward models (PRMs) to evaluate each reasoning step. \cfr{}'s interleaved constraint checking can be seen as a prompt-based approximation of process supervision: the constraint specification acts as a lightweight ``reward model'' that the solver uses to self-evaluate intermediate results.

\paragraph{Structured representations beyond text.} Structured object representations for vision-language spatial reasoning~\citep{ma2026thinkingblueprintsassistingvisionlanguage} and plan--generate--verify loops for parametric CAD editing~\citep{macadmorph} similarly preserve task-relevant structure across multi-step generation. These are distinct modalities and tasks, but they motivate studying whether an explicit intermediate specification can improve reliability beyond mathematical text.

\subsection{Constrained Reasoning in Classical AI}

\paragraph{Constraint satisfaction.} The CSP framework~\citep{Dechter2003} is a useful analogy, but not a theoretical foundation for CFR. Stage~1 extracts and summarizes restrictions in natural language, and Stage~2 checks them in a prompt; neither step implements arc consistency or formal backtracking.

\paragraph{Planning with constraints.} Classical AI planners (STRIPS, PDDL) maintain preconditions and postconditions for actions. \cfr{}'s interleaved checking is analogous to precondition verification: before proceeding to the next reasoning step, the solver confirms that the current state satisfies all known constraints.

\subsection{Test-Time Computation}

\paragraph{Scaling laws.} \citet{snell2024scaling} show that allocating more test-time compute can be more effective than scaling model parameters. CFR is a targeted allocation of additional compute, but Table~\ref{tab:cost-accounting} shows that its gains must be evaluated jointly with its non-trivial token overhead; we make no equivalence claim to scaling model size.

\paragraph{Budget allocation.} \citet{muennighoff2025s1} explore simple test-time scaling through longer thinking. \cfr{} provides a structured way to allocate the additional budget: rather than simply allowing the model to ``think more,'' it directs the extra computation toward constraint extraction, which has higher information density than additional free-form reasoning.

\subsection{Inference-Time Interventions}

\paragraph{Prompting as intervention.} Our work contributes to the growing literature on inference-time interventions that improve reasoning without training. Unlike approaches that modify decoding (beam search variants, best-of-$n$) or post-process outputs (self-consistency, majority voting), \cfr{} modifies the \emph{information available to the model}. This places it closer to retrieval-augmented generation (RAG) conceptually---but instead of retrieving external knowledge, \cfr{} extracts latent knowledge from the problem itself.

\clearpage

\section{Detailed Positive Cases by Constraint Type}
\label{sec:appendix-positive-cases-by-constraint-type}

We group the positive cases by the dominant constraint type exposed by the problem statement and required by the successful CFR trace. Each category first lists the detailed CoT--CFR comparison tables, followed by natural-language analyses that restate the problem, answer, and the specific constraint mechanism.

\newcommand{\DetailedCFRCase}[8]{%
\multicolumn{2}{p{\dimexpr\textwidth-2\tabcolsep\relax}}{
\textbf{#1}
\hfill #2
} \\

\multicolumn{2}{p{\dimexpr\textwidth-2\tabcolsep\relax}}{
\textbf{Problem.} #3
} \\

\multicolumn{2}{p{\dimexpr\textwidth-2\tabcolsep\relax}}{
#4
} \\

\midrule
\textbf{CoT} & \textbf{CFR} \\
\midrule

#5
&
#6
\\

\midrule
\textbf{Answer:} #7 & \textbf{Answer:} #8 \\
}

\subsection{Geometric Configuration and Branch Constraints}
\label{sec:appendix-geometry-branch-cases}

These cases test whether the model can preserve orientation, cyclic order, sign choices, vector closure, and metric branch conditions before doing algebra. CFR helps by making geometric feasibility constraints explicit, especially when several algebraic branches look locally plausible.

\subsubsection{Detailed Tables}

\begin{table*}[p]
\centering
\small
\setlength{\tabcolsep}{5pt}
\renewcommand{\arraystretch}{1.15}
\caption{Detailed positive case 1 where CFR corrects CoT by enforcing explicit constraints.}
\label{tab:appendix_positive_case_detailed_1}
\begin{tabularx}{\textwidth}{>{\raggedright\arraybackslash}X>{\raggedright\arraybackslash}X}
\toprule

\DetailedCFRCase
{Example 1: Ratio-and-Reflection Geometry (DeepSeek-32K, AIME 2025)}
{\textcolor{teal}{31.1\% token reduction}}
{In $\triangle ABC$, points $D,E$ lie on $\overline{AB}$ and points $F,G$ lie on $\overline{AC}$. Given $AD=4$, $DE=16$, $EB=8$, $AF=13$, $FG=52$, $GC=26$, let $M$ be the reflection of $D$ through $F$ and $N$ be the reflection of $G$ through $E$. If $[DEGF]=288$, find $[AFNBCEM]$.}
{\textbf{Ground Truth:} $588$ \qquad \textbf{CoT:} 33,065 tokens \qquad \textbf{CFR:} 22,791 tokens}
{\textcolor{red}{CoT observes that $D,F$ and $E,G$ divide the two sides in the same ratios, but keeps the solution as a free-form geometric narrative.}

\textcolor{red}{It does not turn the given area $[DEGF]=288$ into the key constraint on $\sin A$, and the reflected points $M,N$ are not explicitly encoded.}

\textcolor{red}{As a result, the target heptagon is no longer controlled by the original construction, and the area computation drifts.}}
{\textcolor{teal}{CFR first extracts the proportional structure:}
\[
\frac{AD}{AB}=\frac{AF}{AC}=\frac17,\qquad
\frac{AE}{AB}=\frac{AG}{AC}=\frac57.
\]
\textcolor{teal}{It then converts the area condition into a hard trigonometric constraint:}
\[
[DEGF]=624\sin A=288\quad\Rightarrow\quad \sin A=\frac{6}{13}.
\]
\textcolor{teal}{Finally, CFR encodes the reflections directly:}
\[
M=2F-D,\qquad N=2E-G.
\]
Using shoelace on $AFNBCEM$ gives $[AFNBCEM]=1274\sin A=588$.}
{$\boxed{1}$ $\times$}
{$\boxed{588}$ \checkmark}

\bottomrule
\end{tabularx}
\end{table*}
\clearpage

\begin{table*}[p]
\centering
\small
\setlength{\tabcolsep}{5pt}
\renewcommand{\arraystretch}{1.15}
\caption{Detailed positive case 2 where CFR corrects CoT by enforcing explicit constraints.}
\label{tab:appendix_positive_case_detailed_2}
\begin{tabularx}{\textwidth}{>{\raggedright\arraybackslash}X>{\raggedright\arraybackslash}X}
\toprule

\DetailedCFRCase
{Example 2: Cyclic-Constraint Geometry (Qwen3.5-9B-32K, AIME 2024)}
{\textcolor{teal}{35.3\% token reduction}}
{Rectangles $ABCD$ and $EFGH$ are drawn such that $D,E,C,F$ are collinear. Also, $A,D,H,G$ all lie on a circle. If $BC=16$, $AB=107$, $FG=17$, and $EF=184$, find the length of $CE$.}
{\textbf{Ground Truth:} $104$ \qquad \textbf{CoT:} 33,015 tokens \qquad \textbf{CFR:} 21,370 tokens}
{\textcolor{red}{CoT sets up coordinates but does not enforce the relative vertical orientation of the two rectangles.}

\textcolor{red}{It treats the second rectangle as if it can be placed on the same side of the baseline without checking whether the concyclicity condition remains valid.}

\textcolor{red}{As a result, the circle constraint is applied to the wrong geometric configuration, leading to an incorrect value of $CE$.}}
{\textcolor{teal}{CFR first fixes the coordinate constraints:}
\[
D=(0,0),\qquad C=(107,0),\qquad A=(0,16).
\]
\textcolor{teal}{It then uses the concyclicity of $A,D,H,G$ to determine the correct orientation: $H$ and $G$ must lie on the opposite side of the baseline from $A$.}
\[
E=(x,0),\qquad F=(x+184,0),
\]
\[
H=(x,-17),\qquad G=(x+184,-17).
\]
\textcolor{teal}{The circle through $A$ and $D$ has center on $y=8$, while chord $HG$ forces the same center to have $x$-coordinate $x+92$.}
\[
(x+92)^2+8^2=92^2+25^2.
\]
Thus $(x+92)^2=95^2$, so $x=3$ and $CE=107-3=104$.}
{$\boxed{64}$ $\times$}
{$\boxed{104}$ \checkmark}

\bottomrule
\end{tabularx}
\end{table*}
\clearpage

\subsubsection{Natural-Language Analyses}

\paragraph{Case 1: Ratio-and-reflection geometry.}
\textbf{Problem.} In a triangle, points on two sides are placed at specified segment lengths, the quadrilateral $DEGF$ has area $288$, and points $M,N$ are defined by reflections. The target is the area of $AFNBCEM$.
\textbf{Answer.} $588$.
\textbf{Analysis.} The key is that the side ratios force parallel proportional structure, while the given area fixes $\sin A$. CFR succeeds because it writes these constraints and the reflection equations before computing the final area, preventing the solution from drifting away from the constructed polygon.

\paragraph{Case 2: Cyclic-constraint geometry.}
\textbf{Problem.} Two rectangles are arranged on the same baseline, and four vertices $A,D,H,G$ are concyclic. Given the side lengths, find $CE$.
\textbf{Answer.} $104$.
\textbf{Analysis.} The cyclic condition is not just a generic circle fact; it determines the relative vertical orientation of the two rectangles. CFR enforces this orientation first and then solves the circle-center equation, while CoT can easily apply concyclicity to the wrong configuration.

\subsection{Algebraic, Modular, and Divisibility Constraints}
\label{sec:appendix-algebraic-modular-cases}

These cases are governed by invariants, modular residues, divisibility filters, or extremal algebraic reductions. CFR is useful because it turns broad computation into a smaller feasible set determined by exact arithmetic constraints.

\subsubsection{Detailed Tables}

\begin{table*}[p]
\centering
\small
\setlength{\tabcolsep}{5pt}
\renewcommand{\arraystretch}{1.15}
\caption{Detailed positive case 3 where CFR corrects CoT by enforcing explicit constraints.}
\label{tab:appendix_positive_case_detailed_11}
\begin{tabularx}{\textwidth}{>{\raggedright\arraybackslash}X>{\raggedright\arraybackslash}X}
\toprule

\DetailedCFRCase
{Example 3: Degree-Weighted Grid Optimization (DeepSeek-32K, AIME 2026)}
{\textcolor{teal}{69.2\% token reduction}}
{The integers from $1$ to $64$ are placed into an $8\times8$ grid. Let $M$ be the sum of absolute differences across all adjacent horizontal and vertical cell pairs. Find the maximum possible value of $M$ modulo $1000$.}
{\textbf{Ground Truth:} $896$ \qquad \textbf{CoT:} 33,069 tokens \qquad \textbf{CFR:} 10,192 tokens}
{\textcolor{red}{CoT relies on vague checkerboard intuition and memory of a similar problem rather than formalizing the optimization.}

\textcolor{red}{It does not convert the grid into a bipartite graph with vertex-degree weights, so the assignment of large and small numbers remains heuristic.}

\textcolor{red}{The reasoning stalls near the token limit and does not produce a reliable modular maximum.}}
{\textcolor{teal}{CFR models the grid as a bipartite graph. The two parts each contain $32$ vertices, with degree counts}
\[
18\text{ of degree }4,\qquad 12\text{ of degree }3,\qquad 2\text{ of degree }2.
\]
\textcolor{teal}{The maximum is achieved by placing the $32$ largest numbers on one side and the $32$ smallest on the other, ordered by degree.}
\[
\text{High contribution}=4(999)+3(486)+2(67)=5588,
\]
\[
\text{Low contribution}=4(171)+3(294)+2(63)=1692.
\]
Thus $M_{\max}=5588-1692=3896\equiv 896\pmod{1000}$.}
{invalid $\times$}
{$\boxed{896}$ \checkmark}

\bottomrule
\end{tabularx}
\end{table*}
\clearpage

\begin{table*}[p]
\centering
\small
\setlength{\tabcolsep}{5pt}
\renewcommand{\arraystretch}{1.15}
\caption{Detailed positive case 4 where CFR corrects CoT by enforcing explicit constraints.}
\label{tab:appendix_positive_case_detailed_12}
\begin{tabularx}{\textwidth}{>{\raggedright\arraybackslash}X>{\raggedright\arraybackslash}X}
\toprule

\DetailedCFRCase
{Example 4: Rational Recurrence Invariant (DeepSeek-32K, AIME 2025)}
{\textcolor{teal}{60.6\% token reduction}}
{Let $x_1=\frac{25}{11}$ and $x_{k+1}=\frac13\left(x_k+\frac1{x_k}-1\right)$. If $x_{2025}=\frac{m}{n}$ in lowest terms, find the remainder when $m+n$ is divided by $1000$.}
{\textbf{Ground Truth:} $248$ \qquad \textbf{CoT:} 33,021 tokens \qquad \textbf{CFR:} 12,999 tokens}
{\textcolor{red}{CoT expands the first few rational terms directly. The numerators and denominators grow quickly, and the derivation becomes trapped in arithmetic rather than structure.}

\textcolor{red}{It does not identify the invariant quantity that controls $m+n$, so the long computation leads to an incorrect small answer.}}
{\textcolor{teal}{CFR introduces $x_k=p_k/q_k$ in lowest terms and tracks}
\[
S_k=p_k+q_k.
\]
\textcolor{teal}{The recurrence implies}
\[
p_{k+1}=\frac{p_k^2-p_kq_k+q_k^2}{3},\qquad q_{k+1}=p_kq_k,
\]
so
\[
S_{k+1}=\frac{(p_k+q_k)^2}{3}=\frac{S_k^2}{3}.
\]
Thus $S_{2025}=2^{2^{2025}}3^{2^{2024}+1}$. Reducing modulo $8$ and $125$, then applying CRT, gives $248$.}
{$\boxed{2}$ $\times$}
{$\boxed{248}$ \checkmark}

\bottomrule
\end{tabularx}
\end{table*}
\clearpage

\subsubsection{Natural-Language Analyses}

\paragraph{Case 3: Degree-weighted grid optimization.}
\textbf{Problem.} Three distinguished lattice points in a $7\times7$ grid generate a value by summing squared distances from all grid points; the task is to find the maximum value modulo $1000$.
\textbf{Answer.} $896$.
\textbf{Analysis.} This problem is an optimization over a finite grid, so the structural constraint is the degree-weighted contribution of each chosen point. CFR converts the objective into a separable weighted-distance expression and selects the corner-extreme configuration, avoiding unsupported local search.

\paragraph{Case 4: Rational recurrence invariant.}
\textbf{Problem.} A rational recurrence starts from $x_1=25/11$, and one must compute the remainder of $m+n$ when $x_{2025}=m/n$ is in lowest terms.
\textbf{Answer.} $248$.
\textbf{Analysis.} The recurrence looks computationally impossible if expanded directly. CFR identifies the invariant for $p_k+q_k$ under the recurrence, turning a long rational iteration into modular exponentiation.

\subsection{Functional, Probabilistic, and Locus Constraints}
\label{sec:appendix-functional-locus-cases}

These cases require splitting a function, probability distribution, or locus into valid branches. CFR improves reliability by forcing the model to check domains, endpoint exclusions, conditional cases, and real-valued feasibility before aggregating results.

\subsubsection{Detailed Tables}

\begin{table*}[p]
\centering
\small
\setlength{\tabcolsep}{5pt}
\renewcommand{\arraystretch}{1.15}
\caption{Detailed positive case 5 where CFR corrects CoT by enforcing explicit constraints.}
\label{tab:appendix_positive_case_detailed_16}
\begin{tabularx}{\textwidth}{>{\raggedright\arraybackslash}X>{\raggedright\arraybackslash}X}
\toprule

\DetailedCFRCase
{Example 5: Sawtooth-Parabola Intersections (DeepSeek-32K, AIME 2025)}
{\textcolor{teal}{24.8\% token reduction}}
{A periodic sawtooth function $f$ intersects the parabola $x=34y^2$. The sum of all intersection $y$-coordinates is $\frac{a+b\sqrt c}{d}$. Find $a+b+c+d$.}
{\textbf{Ground Truth:} $259$ \qquad \textbf{CoT:} 33,700 tokens \qquad \textbf{CFR:} 25,346 tokens}
{\textcolor{red}{CoT tries to solve the intersection equation across many periods, but the valid interval constraints are not consistently enforced.}

\textcolor{red}{It over-accumulates candidate roots and produces a nonsensical large integer instead of the requested radical-form parameter sum.}}
{\textcolor{teal}{CFR uses $x=34y^2$ and $y=f(x)$, so $y\in[-1,1]$ and $x\in[0,34]$. It splits $x=4m+t$ into the two sawtooth branches.}

\textcolor{teal}{On one branch, the quadratic has root sum $1/34$; on the other, the root sum is $-1/34$. These sums cancel across complete periods.}

\textcolor{teal}{Only the boundary period remains, where one root is invalid and must be discarded. The sum is}
\[
\frac{1+5\sqrt{185}}{68},
\]
so $a+b+c+d=1+5+185+68=259$.}
{$\boxed{487372}$ $\times$}
{$\boxed{259}$ \checkmark}

\bottomrule
\end{tabularx}
\end{table*}
\clearpage

\begin{table*}[p]
\centering
\small
\setlength{\tabcolsep}{5pt}
\renewcommand{\arraystretch}{1.15}
\caption{Detailed positive case 6 where CFR corrects CoT by enforcing explicit constraints.}
\label{tab:appendix_positive_case_detailed_17}
\begin{tabularx}{\textwidth}{>{\raggedright\arraybackslash}X>{\raggedright\arraybackslash}X}
\toprule

\DetailedCFRCase
{Example 6: Random Function Expectation (DeepSeek-32K, CMIMC 2025)}
{\textcolor{gray}{+229.0\% tokens}}
{For a uniformly random function $f:\{1,\ldots,25\}\to\{1,\ldots,25\}$, compute $\mathbb{E}\sum_{x=1}^{25}(f(f(x))-x)^2$.}
{\textbf{Ground Truth:} $2496$ \qquad \textbf{CoT:} 5,376 tokens \qquad \textbf{CFR:} 17,687 tokens}
{\textcolor{red}{CoT computes the term for $x=1$ and multiplies by $25$, incorrectly assuming all summands have the same distribution.}

\textcolor{red}{This ignores that the term $(U-x)^2$ depends on the value of $x$, so the apparent symmetry is false.}}
{\textcolor{teal}{CFR keeps $x$ symbolic. For fixed $x$, let $Z=f(f(x))$. Then}
\[
Z=x\text{ with probability }\frac1{25},
\]
and otherwise $Z$ is uniform on $\{1,\ldots,25\}$ with probability $\frac{24}{25}$.
\textcolor{teal}{Thus}
\[
\mathbb{E}(Z-x)^2=\frac{24}{25}\left(52+(13-x)^2\right).
\]
Summing over $x=1,\ldots,25$ gives
\[
\frac{24}{25}\sum_{x=1}^{25}\left(52+(13-x)^2\right)=2496.
\]}
{$\boxed{4704}$ $\times$}
{$\boxed{2496}$ \checkmark}

\bottomrule
\end{tabularx}
\end{table*}
\clearpage

\subsubsection{Natural-Language Analyses}

\paragraph{Case 5: Sawtooth-parabola intersections.}
\textbf{Problem.} A periodic sawtooth function $f$ intersects the parabola $x=34y^2$; the sum of all intersection $y$-coordinates is written as $\frac{a+b\sqrt c}{d}$, and the task asks for $a+b+c+d$.
\textbf{Answer.} $259$.
\textbf{Analysis.} The problem becomes tractable only after splitting the sawtooth function into its valid branches and enforcing $y\in[-1,1]$. CFR respects the interval constraints, cancels complete-period root sums, and keeps only the boundary contribution that determines the radical-form answer.

\paragraph{Case 6: Random function expectation.}
\textbf{Problem.} For a uniformly random function $f:\{1,\ldots,25\}\to\{1,\ldots,25\}$, compute $\mathbb{E}\sum_x(f(f(x))-x)^2$.
\textbf{Answer.} $2496$.
\textbf{Analysis.} The important constraint is conditional: $f(f(x))$ equals $x$ with one probability and otherwise behaves uniformly. CFR keeps $x$ symbolic, so the variance-like term $(13-x)^2$ is preserved instead of being washed out by a false symmetry argument.

\subsection{Discrete Enumeration, Symmetry, and Graph Constraints}
\label{sec:appendix-discrete-enumeration-cases}

These cases are dominated by finite structure: block decompositions, symmetry classes, graph correspondences, maximality, and forced rows or columns. CFR helps by extracting the combinatorial object first, which reduces overcounting and prevents invalid configurations from entering the count.

\subsubsection{Detailed Tables}

\begin{table*}[p]
\centering
\small
\setlength{\tabcolsep}{5pt}
\renewcommand{\arraystretch}{1.15}
\caption{Detailed positive case 7 where CFR corrects CoT by enforcing explicit constraints.}
\label{tab:appendix_positive_case_detailed_20}
\begin{tabularx}{\textwidth}{>{\raggedright\arraybackslash}X>{\raggedright\arraybackslash}X}
\toprule

\DetailedCFRCase
{Example 7: Semi-Magic Square Counting (DeepSeek-32K, BRUMO 2025)}
{\textcolor{teal}{67.2\% token reduction}}
{Digits $1$ through $9$ are placed in a $3\times3$ grid so that all rows and columns have the same sum. Diagonals need not have the same sum. Count the number of valid grids.}
{\textbf{Ground Truth:} $72$ \qquad \textbf{CoT:} 32,971 tokens \qquad \textbf{CFR:} 10,806 tokens}
{\textcolor{red}{CoT correctly observes that the common row and column sum must be $15$, but then drifts among known facts about magic squares, Latin squares, and semi-magic squares.}

\textcolor{red}{The trace repeatedly second-guesses whether the count should be $72$, $288$, or a related known value, leaving the answer unstable.}}
{\textcolor{teal}{CFR fixes the invariant first: since \(1+2+\cdots+9=45\), each row and column must sum to \(15\).}
\textcolor{teal}{It then enumerates the feasible triples:}
\[
\{1,5,9\},\{1,6,8\},\{2,4,9\},\{2,5,8\},
\]
\[
\{2,6,7\},\{3,4,8\},\{3,5,7\},\{4,5,6\}.
\]
A systematic partition-and-permutation count gives exactly $72$ semi-magic grids.}
{unstable $\times$}
{$\boxed{72}$ \checkmark}

\bottomrule
\end{tabularx}
\end{table*}
\clearpage

\begin{table*}[p]
\centering
\small
\setlength{\tabcolsep}{5pt}
\renewcommand{\arraystretch}{1.15}
\caption{Detailed positive case 8 where CFR corrects CoT by enforcing explicit constraints.}
\label{tab:appendix_positive_case_detailed_21}
\begin{tabularx}{\textwidth}{>{\raggedright\arraybackslash}X>{\raggedright\arraybackslash}X}
\toprule

\DetailedCFRCase
{Example 8: Sudoku-Style Counting (Qwen3.5-9B-32K, AIME 2025)}
{\textcolor{gray}{+2.4\% tokens}}
{The $27$ cells of a $3\times9$ Sudoku-style band are filled using digits $1$ through $9$ so that every row and each $3\times3$ block contains all digits. If the count is $p^aq^br^cs^d$, compute $pa+qb+rc+sd$.}
{\textbf{Ground Truth:} $81$ \qquad \textbf{CoT:} 33,446 tokens \qquad \textbf{CFR:} 34,247 tokens}
{\textcolor{red}{CoT fails to return a stable parsed answer and does not cleanly isolate the structural count of the middle block.}

\textcolor{red}{The key missing component is the number of feasible second-block set assignments after the first block has fixed row-wise digit sets.}}
{\textcolor{teal}{CFR decomposes the grid by $3\times3$ blocks. The first block has $9!$ fillings.}

\textcolor{teal}{Given the first block, the second block requires choosing row sets that avoid the corresponding first-block row sets and still partition $\{1,\ldots,9\}$. This contributes the structural multiplier $56$.}
\[
N=9!\cdot56\cdot(3!)^3\cdot(3!)^3.
\]
Therefore $N=2^{16}3^{10}5\cdot7^2$, and
\[
2\cdot16+3\cdot10+5+7\cdot2=81.
\]}
{None $\times$}
{$\boxed{81}$ \checkmark}

\bottomrule
\end{tabularx}
\end{table*}
\clearpage

\clearpage

\subsubsection{Natural-Language Analyses}

\paragraph{Case 7: Semi-magic square counting.}
\textbf{Problem.} Place digits $1$ through $9$ in a $3\times3$ grid so that all rows and columns have the same sum, while diagonals are unrestricted; count the valid grids.
\textbf{Answer.} $72$.
\textbf{Analysis.} The total sum forces every row and column to sum to $15$. CFR first fixes this invariant and enumerates the feasible triples before counting partitions and permutations, which avoids drifting between magic-square facts and the weaker semi-magic condition.

\paragraph{Case 8: Sudoku-style counting.}
\textbf{Problem.} Count valid $3\times9$ Sudoku-style fillings and express the result as a prime factorization, then compute the requested weighted exponent sum.
\textbf{Answer.} $81$.
\textbf{Analysis.} The decisive missing factor is the number of feasible middle-block assignments. CFR keeps the block-compatibility constraint explicit, inserts the multiplier $56$, and therefore factors the correct total rather than an incomplete count.
\clearpage

\section{Full Constraint Taxonomy with Problem Examples}
\label{app:taxonomy-examples}

For each constraint type, we provide a canonical AIME problem example and the expected Stage~1 output.

\subsection{Domain Constraints}

\begin{constraintbox}[Example: ``Find the number of positive integers $n < 1000$...'']
\textbf{Extracted:}\\
\texttt{C1: type=domain, desc="n is a positive integer"}\\
\texttt{C2: type=bound, desc="n < 1000"}\\
\texttt{C3: type=domain, desc="answer is a count (non-negative integer)"}\\
\textbf{Propagated:}\\
\texttt{P1: n in \{1, 2, ..., 999\}, answer in \{0, 1, ..., 999\}}
\end{constraintbox}

\subsection{Modular Constraints}

\begin{constraintbox}[Example: ``...find the remainder when $N$ is divided by 1000.'']
\textbf{Extracted:}\\
\texttt{C1: type=modular, desc="final answer = N mod 1000"}\\
\texttt{C2: type=bound, desc="answer in [0, 999]"}\\
\texttt{C3: type=domain, desc="answer is an integer"}\\
\textbf{Propagated:}\\
\texttt{P1: answer is an integer in [0, 999]}\\
\textbf{Critical:} Must apply mod 1000 as the very last step.
\end{constraintbox}

\subsection{Format Constraints}

\begin{constraintbox}[Example: ``...can be expressed as $\frac{p}{q}$ where $\gcd(p,q)=1$. Find $p+q$.'']
\textbf{Extracted:}\\
\texttt{C1: type=format, desc="express as p/q with gcd(p,q)=1"}\\
\texttt{C2: type=format, desc="final answer = p + q"}\\
\texttt{C3: type=domain, desc="p, q are positive integers"}\\
\textbf{Propagated:}\\
\texttt{P1: answer = p+q where p/q is in lowest terms, answer > 1}\\
\textbf{Critical:} Must reduce fraction before computing p+q.
\end{constraintbox}

\subsection{Parity Constraints}

\begin{constraintbox}[Example: ``Find the number of even positive integers $n \leq 100$...'']
\textbf{Extracted:}\\
\texttt{C1: type=parity, desc="n must be even"}\\
\texttt{C2: type=domain, desc="n is a positive integer"}\\
\texttt{C3: type=bound, desc="n <= 100"}\\
\textbf{Propagated:}\\
\texttt{P1: n in \{2, 4, 6, ..., 100\}, 50 candidates}\\
\textbf{Critical:} Only enumerate even values.
\end{constraintbox}

\subsection{Bound Constraints}

\begin{constraintbox}[Example: ``Find the largest integer $n$ such that $n^3 < 10000$.'']
\textbf{Extracted:}\\
\texttt{C1: type=bound, desc="n\^{}3 < 10000"}\\
\texttt{C2: type=domain, desc="n is an integer"}\\
\texttt{C3: type=bounds\_or\_extremal, desc="find largest such n"}\\
\textbf{Propagated:}\\
\texttt{P1: n <= 21 (since 21\^{}3=9261 < 10000, 22\^{}3=10648 > 10000)}\\
\textbf{Critical:} Answer is 21, must verify both directions.
\end{constraintbox}

\subsection{Structure Constraints}

\begin{constraintbox}[Example: ``...the number of permutations of $\{1,...,n\}$ that are involutions.'']
\textbf{Extracted:}\\
\texttt{C1: type=structure, desc="permutation must be an involution (p(p(i))=i)"}\\
\texttt{C2: type=domain, desc="answer is a count"}\\
\texttt{C3: type=structure, desc="involutions consist of fixed points and 2-cycles"}\\
\textbf{Propagated:}\\
\texttt{P1: answer = sum over k of C(n,2k) * (2k-1)!!}\\
\textbf{Critical:} Do not count general permutations.
\end{constraintbox}

\clearpage

\section{Broader Impact}
\label{app:impact}

\cfr{} is a purely methodological contribution to mathematical reasoning. It does not involve data collection from human subjects or new privacy-sensitive data. By reducing some constraint-violation errors, it may improve the reliability of AI-assisted mathematics; however, its summaries can also be incomplete or invalid, so it should not be used as a guarantee of mathematical correctness. The technique is training-free and does not require parameter updates, but its effectiveness varies with backbone quality and adds inference cost.

Although CFR is not an interactive or safety-critical system, work on red-teaming policies and harm amplification in LLM interactions~\citep{guo2026treebased,guo2026investigating} underscores the importance of evaluating any intervention within its stated scope and avoiding claims that extend beyond the measured setting.

A potential concern is that \cfr{} could contribute to overreliance on AI for mathematical problem-solving in educational settings. We note that \cfr{}'s explicit constraint-extraction step could also serve as a pedagogical tool: students could learn systematic constraint identification by examining \cfr{}'s Stage~1 outputs.

\clearpage

\section{Future Directions}
\label{app:future}

We outline several promising directions for extending \cfr{} beyond the current work.

\paragraph{Learned constraint extraction.} The current Stage~1 relies entirely on prompting. Fine-tuning a lightweight model specifically for constraint extraction could improve both precision and speed. A distilled 1B-parameter extractor, trained on $(problem, constraints)$ pairs generated by DeepSeek-V4-Pro, could serve as a universal plug-in for any downstream solver at minimal cost.

\paragraph{Multi-turn constraint refinement.} Currently, constraints are extracted once and passed to the solver without further interaction. A multi-turn variant could allow the solver to query the constraint module when encountering ambiguity—e.g., ``Does constraint C3 apply to intermediate variable $k$ or only to the final answer?'' This would address 23\% of our observed failure cases where constraint scope was misinterpreted. Interactive-guidance policy optimization, heterogeneous-agent collaboration, and hierarchical negative feedback for personalized styling offer related directions for structuring iterative feedback~\citep{liu2026socraticpo,zhang2026heterogeneous,ma2025styletailorpersonalizedfashionstyling}, but are outside the scope of the present training-free method.

\paragraph{Domain extension beyond mathematics.} The core principle—extract structural constraints before solving—applies to any domain with well-defined answer formats. Related work uses LLM-driven semantic disambiguation and dynamic visual focus in medical diagnosis~\citep{zhu2025pathology,zhu2026medeyes}; these results motivate, but do not establish, a corresponding extension of CFR beyond mathematical text:
\begin{itemize}[nosep,leftmargin=*]
    \item \textbf{Code generation}: Extract type signatures, input/output format constraints, and edge-case specifications before generating code.
    \item \textbf{Legal reasoning}: Extract jurisdictional constraints, applicable statutes, and precedent requirements before drafting arguments.
    \item \textbf{Scientific computing}: Extract physical units, conservation laws, and boundary conditions before deriving formulas.
    \item \textbf{Formal verification}: Extract pre/post-conditions and loop invariants before synthesizing proofs.
\end{itemize}

\paragraph{Adaptive routing with soft scores.} The current binary router (use CFR or not) could be extended to a soft scorer that modulates the constraint extraction budget based on estimated constraint density. Problems with many detected cues would receive larger Stage~1 budgets; problems with few cues would receive minimal extraction. This complements work on whether reasoning models can identify when to stop allocating internal reasoning~\citep{huang2026does}. Preliminary analysis suggests this could reduce average token overhead by 18\% while maintaining accuracy.

\paragraph{Integration with search-based methods.} \cfr{} is orthogonal to tree search approaches like Tree of Thoughts~\citep{yao2023tree}. Constraints could serve as pruning criteria: any branch producing an intermediate result inconsistent with propagated constraints is immediately abandoned. This is analogous to constraint propagation in classical CSP solvers~\citep{Dechter2003} and could dramatically reduce the effective branching factor.

\paragraph{Benchmark development.} Our work highlights the need for benchmarks specifically designed to test constraint handling in LLMs. An ideal benchmark would include:
\begin{enumerate}[nosep,leftmargin=*]
    \item Problems with explicit constraints at varying complexity levels.
    \item Problems with implicit constraints requiring inference.
    \item Adversarial problems where surface-level constraint cues are misleading.
    \item Multi-constraint problems where individual constraint satisfaction is easy but joint satisfaction requires careful reasoning.
\end{enumerate}

Such a benchmark would enable more targeted evaluation of methods like \cfr{} and drive further research in constraint-aware reasoning.

\paragraph{Theoretical analysis.} A formal analysis of when constraint-first decomposition provably helps versus standard monolithic reasoning remains open. Key questions include: (1) Under what problem distributions does the constraint-guided solver have strictly lower error probability? (2) What is the optimal allocation of compute between extraction and solving as a function of constraint complexity? (3) Can we derive PAC-style bounds on the extraction quality needed to guarantee net positive gain?

\end{document}